\documentclass[journal]{IEEEtran}
\usepackage{amsmath,amsfonts}
\usepackage{algorithm}
\usepackage{array}
\usepackage{booktabs}
\usepackage{multirow}
\usepackage[caption=false,font=normalsize,labelfont=sf,textfont=sf]{subfig}
\usepackage{textcomp}
\usepackage{stfloats}
\usepackage{url}
\usepackage{verbatim}
\usepackage{graphicx}
\usepackage{cite}
\usepackage{color}
\usepackage{hyperref}
\hypersetup{hidelinks, colorlinks=True, allcolors=magenta, pdfstartview=Fit, breaklinks=true}

\usepackage{balance}

\begin{document}

\title{Efficient Fourier Filtering Network with Contrastive Learning for AAV-based Unaligned Bimodal Salient Object Detection}

\author{Pengfei Lyu, Pak-Hei Yeung, Xiaosheng Yu, Xiufei Cheng, Chengdong Wu, 

and Jagath C. Rajapakse, \IEEEmembership{Fellow, IEEE}

\thanks{This work was supported in part by the National Natural Science Foundation of China under Grant 62403108 and 62306187, the Foundation of Ministry of Industry and Information Technology TC220H05X-04, the Liaoning Provincial Natural Science Foundation Joint Fund 2023-MSBA-075, the Fundamental Research Fund for the Central Universities of China under Grant N2426005 and N2326001, and the China Scholarship Fund. \textit{(Corresponding author: Chengdong Wu and Jagath C. Rajapakse).}

Pengfei Lyu is with the Faculty of Robot Science and Engineering, Northeastern University, Shenyang, 110169 China, and with the College of Computing and Data Science, Nanyang Technological University, 639798 Singapore (e-mail: lyupengfei@stumail.neu.edu.cn).

Pak-Hei Yeung and Jagath C. Rajapakse are with the College of Computing and Data Science, Nanyang Technological University, 639798 Singapore (e-mail: pakhei.yeung@ntu.edu.sg; asjagath@ntu.edu.sg).

Xiaosheng Yu, Xiufei Cheng, and Chengdong Wu are with the Faculty of Robot Science and Engineering, Northeastern University, Shenyang, 110169 China (e-mail: yuxiaosheng@mail.neu.edu.cn; chengxiufei@stumail.neu.edu.cn; wuchengdong@mail.neu.edu.cn). 
}}

\markboth{Journal of \LaTeX\ Class Files,~Vol.~18, No.~9, September~2020}%
{How to Use the IEEEtran \LaTeX \ Templates}
\maketitle

\begin{abstract}
Autonomous aerial vehicle (AAV)-based bimodal salient object detection (BSOD) aims to segment salient objects in a scene utilizing complementary cues in unaligned RGB and thermal image pairs. However, the high computational expense of existing AAV-based BSOD models limits their applicability to real-world AAV devices. To address this problem, we propose an efficient Fourier filter network with contrastive learning that achieves both real-time and accurate performance. Specifically, we first design a semantic contrastive alignment loss to align the two modalities at the semantic level, which facilitates mutual refinement in a parameter-free way. Second, inspired by the fast Fourier transform that obtains global relevance in linear complexity, we propose synchronized alignment fusion, which aligns and fuses bimodal features in the channel and spatial dimensions by a hierarchical filtering mechanism. Our proposed model, AlignSal, reduces the number of parameters by 70.0\%, decreases the floating point operations by 49.4\%, and increases the inference speed by 152.5\% compared to the cutting-edge BSOD model (\textit{i.e.}, MROS). Extensive experiments on the AAV RGB-T 2400 and 
%three weakly aligned datasets 
seven bimodal dense prediction datasets
demonstrate that AlignSal achieves both real-time inference speed and better performance and generalizability compared to nineteen state-of-the-art models across most evaluation metrics.
In addition, our ablation studies further verify AlignSal's potential in boosting the performance of existing aligned BSOD models on AAV-based unaligned data. The code is available at: \url{https://github.com/JoshuaLPF/AlignSal}.
\end{abstract}

\section{Introduction}

\begin{figure}[!htp]
	\centering \includegraphics[width=0.43\textwidth]{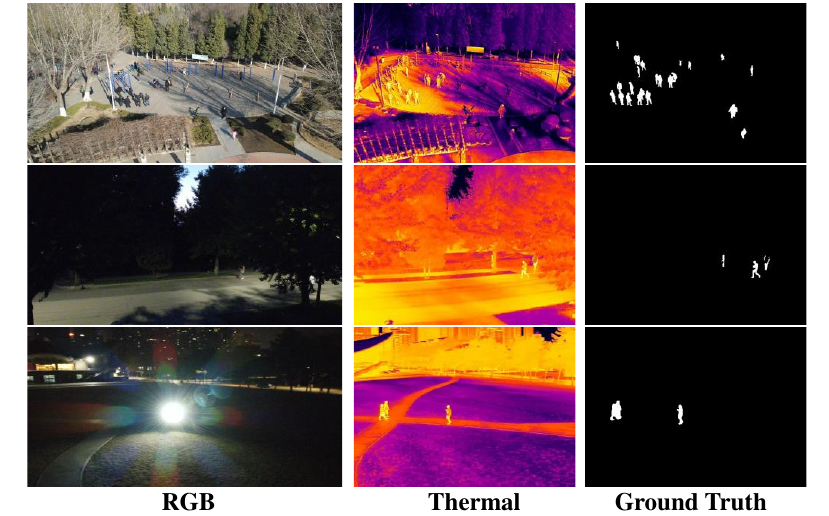}
	\caption{Examples from AAV RGB-T 2400 dataset \cite{10315195}. RGB images typically have higher resolution and a wider field of view compared to thermal images.}
\label{fig:1}
\end{figure}

\IEEEPARstart{A}{utonomous} aerial
vehicle (AAV)-based object detection methods have been broadly applied in various fields, such as emergency rescue \cite{bo2022basnet,xu2024yoloow}, plant maintenance \cite{hu2023detection}, and electrical overhaul \cite{senthilnath2021bs}. 
Despite the impressive achievement of existing detection models \cite{bo2022basnet,zhang2023cfanet,li2020comnet,fang2024scinet}, 
methods that rely solely on RGB \cite{bo2022basnet,zhang2023cfanet} or thermal imaging \cite{li2020comnet,fang2024scinet} face challenges in scenes with low illumination, occlusion, and background clutter. 
The two modalities have their own characteristics.
While the RGB modality provides rich color and texture information about a scene, 
the thermal modality utilizes temperature differences to identify objects independently of lighting conditions. 
Therefore, recent works \cite{zhang2022visible,10315195} have attempted to combine the strengths of RGB and thermal modalities to improve AAV-based object detection in complex scenes.

In practice, the resolution and scale of images captured by the RGB and thermal cameras mounted on AAVs may differ, resulting in unaligned bimodal images, as shown in Fig. \ref{fig:1}. These differences lead to varying proportions and spatial locations of the same object between the two modalities. However, most existing bimodal object detection datasets \cite{peng2014rgbd,9767629,10003255} are manually aligned, avoiding many real-world challenges due to the misalignment between the two modalities. 
Consequently, models designed based on these aligned datasets \cite{9454273,9505635,jin2022moadnet,9611276,9803225,9926193,10003255,sun2023catnet,9869666,10015667,10127616,wang2024learning} usually underperform when applied to unaligned data under realistic conditions. 

To address these issues, Song et al. \cite{10315195} constructed the first unaligned bimodal salient object detection (BSOD) dataset, AAV RGB-T 2400, based on the AAV view and proposed a paradigm model named MROS. 
MROS achieves remarkable performance by aligning the two modalities, then performing modal fusion, followed by decoding and generating predictions. 
However, two primary challenges remain. 
First, AAVs are often used for real-time scene detection, but the complexity of MROS - with 90.4M parameters and 45.3G floating point operations - places a significant burden on the AAV's processor and memory. 
In particular, modal alignment and fusion modules have high computational demands, 
limiting the overall inference speed. 
Second, since AAVs operate at high altitudes, the scale of the objects captured is small, which increases spatial location offsets of small-scale objects between the two modalities. The alignment strategies in MROS, based on convolutional attention operations \cite{woo2018cbam}, struggle to handle large spatial offsets effectively due to their limited receptive fields.

In this paper, we propose a two-step solution to address the aforementioned limitations of the current BSOD methods. 
First, we aim to develop parameter-free learning strategies that reduce model complexity while achieving essential modal alignment. 
Second, we simultaneously perform weak bimodal alignment and fusion, further minimizing testing complexity to improve inference speed. 
By integrating bimodal features into a shared feature space, we aim to establish a multi-dimensional, multi-filtering mechanism based on the mutual relationships between the two modalities on a global scale to address varying offset scales and leverage complementary bimodal information effectively.
To achieve this, in this work, we propose \textbf{AlignSal}, an efficient Fourier filtering network with contrastive learning for accurate and real-time AAV-based unaligned BSOD. 

Specifically, AlignSal is composed of two major novelties that contribute to its superior performance. 
First, we propose the semantic contrastive alignment loss (SCAL) to align RGB and thermal modalities at the semantic level. 
Inspired by contrastive learning \cite{radford2021learning,yang2021taco,he2023align,jiang2024t,song2024contrastalign}, 
SCAL forces similar local features in RGB and thermal modalities to stay close in the embedding space while pushing the dissimilar features apart. 
This knowledge exchange process allows both modalities to refine each other, 
achieving modal alignment while improving their individual representations. 
Notably, SCAL does not increase computational loads during inference.
Second, we propose the synchronized alignment fusion (SAF) module,
which leverages the fast Fourier transform to align features in the channel and spatial dimensions and facilitate bimodal fusion. 
By modeling channel dependencies, SAF employs a multi-group global filtering mechanism that hierarchically captures spatial offsets and bimodal salient cues at multiple scales, 
further aligning and integrating them into the fused feature. 
The low complexity of SAF ensures real-time performance during inference. 
With the proposed SCAL and SAF, AlignSal effectively aligns RGB and thermal modalities by leveraging their correlations and complementary strengths,
producing accurate saliency maps from the AAV perspective.

In summary, our contributions are as follows:
\begin{itemize}
    \item We propose an efficient model, AlignSal, for AAV-based unaligned BSOD. 
    With detailed ablation studies, we validate the effectiveness of the proposed key components, namely SCAL and SAF.
    We further demonstrate SCAL's ability to enhance the performance of existing aligned BSOD models on the unaligned bimodal data.
    \item With extensive experiments conducted on the AAV RGB-T 2400 and 
    %three weakly aligned datasets,
    seven additional bimodal dense prediction datasets,
    we show that our proposed AlignSal achieves better performance and generalizability compared to nineteen existing state-of-the-art models,
    while achieving real-time inference speed.
    \item Compared to the current top-performing model (\textit{i.e.} MROS), 
    AlignSal achieves better and more robust performance in different scenarios across most evaluation metrics,
    while improving the inference speed by 152.5\% (30.8 frames per second\footnote{Tested on an NVIDIA RTX4060 laptop GPU.}) and using only 30.0\% of MROS' parameters (27.1M) and 50.6\% of its floating point operations (22.9G).
\end{itemize}

\section{Related Work}
\subsection{Bimodal Salient Object Detection (BSOD)}
With advancements in sensor technology, bimodal salient object detection (BSOD) has emerged as a prominent research area \cite{wang2021cgfnet,gao2021unified,zhao2022self,zhang2021deep,lee2022spsn,wu2023hidanet}. 
The introduction of thermal information enhances the model's ability to distinguish objects in complex environments \cite{wang2018rgb,shigematsu2017learning,song2022novel}. 
Thermal images capture temperature distributions, highlighting object contours, while RGB images deliver abundant color and contextual information that enables a more intuitive understanding of the scene.
To leverage inter-modal complementary cues, existing BSOD models can be categorized into early, middle, and late fusion approaches.

Early fusion models typically aggregate bimodal image pairs linearly as input \cite{zhao2020single,9803225,fu2021siamese}. While this strategy reduces computational load, it fails to prevent the inter-modal noise from interfering with each other. In contrast, late fusion models often employ dual- \cite{pang2020hierarchical,9454273} or triple-stream branches \cite{ji2021calibrated,cong2022cir} that focus on semantic features or decoder feature fusion.
However, these approaches overlook features at early levels that retain important fine-grained spatial information. 
Therefore, the balanced middle fusion strategy has become the dominant approach in BSOD \cite{song2023potential,zhou2021ccafnet,9505635,jin2022moadnet,9926193,10003255,10042233,10127616,wang2024learning}. 
For instance, Huo et al. \cite{9505635} proposed an ultra-
lightweight network with cross-modal filtering and step-wise refinement,
while a global illumination module was introduced \cite{9926193} to determine the use of the thermal modality. 
Zhou et al. \cite{10127616} developed a wavelet-based model that uses knowledge distillation to transfer information from a complex model. 
Wang et al. \cite{wang2024learning} proposed an adaptive fusion bank, which utilizes basic fusion schemes to address different challenges simultaneously. 
Despite the significant advances in these models, their performance is constrained by the limited view of convolution operations.
Inspired by the attention mechanism employed by the Transformer \cite{vaswani2017attention} which is good at capturing global information, researchers have applied Transformer-based solutions to BSOD \cite{liu2021visual,wu2023transformer,hu2024cross,zhou2023position,9611276,sun2023catnet,cong2023point,9869666,10015667}. Specifically, recent works \cite{hu2024cross,9611276,sun2023catnet,9869666} employed Transformer-based backbones to model initial multi-level features. 
Zhou et al. \cite{zhou2023position} designed an auxiliary task to optimize pixel classification and proposed a Transformer-based decoder. 
Pang et al. \cite{10015667} proposed view-mixed attention for integrating cross-modal and cross-level features. 
Cong et al. \cite{cong2023point} introduced position constraints to enhance cross-modal long-range modeling. 
Although these models effectively address BSOD challenges, they are designed for manually aligned bimodal image pairs and struggle with unaligned instantaneous data.
In contrast, our proposed model, AlignSal, is developed for unaligned BSOD data, which can process unaligned RGB and thermal image pairs in real time and exploit the inter-modal complementarity to accurately identify salient objects.

Recently, Tu et al. \cite{tu2022weakly} simulated weakly aligned data by applying affine transformations to aligned bimodal data and used dynamic convolution for weakly aligned BSOD. However, such artificially unaligned data does not fully capture real-world conditions. 
To address this, Song et al. \cite{10315195} constructed an unaligned BSOD dataset based on AAV views of urban scenes and proposed the MROS model, which performs image alignment followed by fusion. 
While MROS demonstrates excellent performance, its high computational demands hinder real-time detection capabilities.
In order to prioritize both accuracy and real-time performance, we allocate resource-intensive bimodal alignment to the training phase, which shifts part of the computational burden from inference to training.
We also design a hierarchical global filtering mechanism with low complexity to handle modal fusion and alignment. 
Our model achieves competitive performance compared to state-of-the-art models while enabling real-time detection with manageable complexity.

\subsection{Contrastive Learning (CL)}
Contrastive learning (CL) focuses on learning data representations by analyzing the similarities and differences between samples. In CL, the distance between features of similar samples in the embedding space is minimized while that between dissimilar samples is maximized, facilitating the acquisition of meaningful feature representations \cite{hadsell2006dimensionality}. CL has been successfully applied in various fields of computer vision \cite{radford2021learning,wang2021exploring,jiang2024t,sung2024contextrast,wu2023transformer,song2024contrastalign}. 
For instance, Radford et al. \cite{radford2021learning} proposed a model for aligning visual and textual information through CL, and Jiang et al. \cite{jiang2024t} combined visual and textual prompts within a single model via CL. 
Song et al. \cite{song2024contrastalign} aligned RGB and radar features using CL to improve the robustness of fusion. Beyond its use for aligning heterogeneous modalities, CL has also been widely studied to enhance inter-pixel discrimination. 
Wang et al. \cite{wang2021exploring} extended pixel-level CL from a single image to neighboring images. 
Sung et al. \cite{sung2024contextrast} proposed contextual-based CL to obtain clear feature distinctions.
Wu et al. \cite{wu2023transformer} utilized CL to increase the separation between foreground and background pixels in predictions. 
In this paper, to address the misalignment of RGB and thermal modalities, we leverage CL to align bimodal features at the semantic level and address AAV-based unaligned BSOD.

\begin{figure*}[!htp]
	\centering \includegraphics[width=1\textwidth]{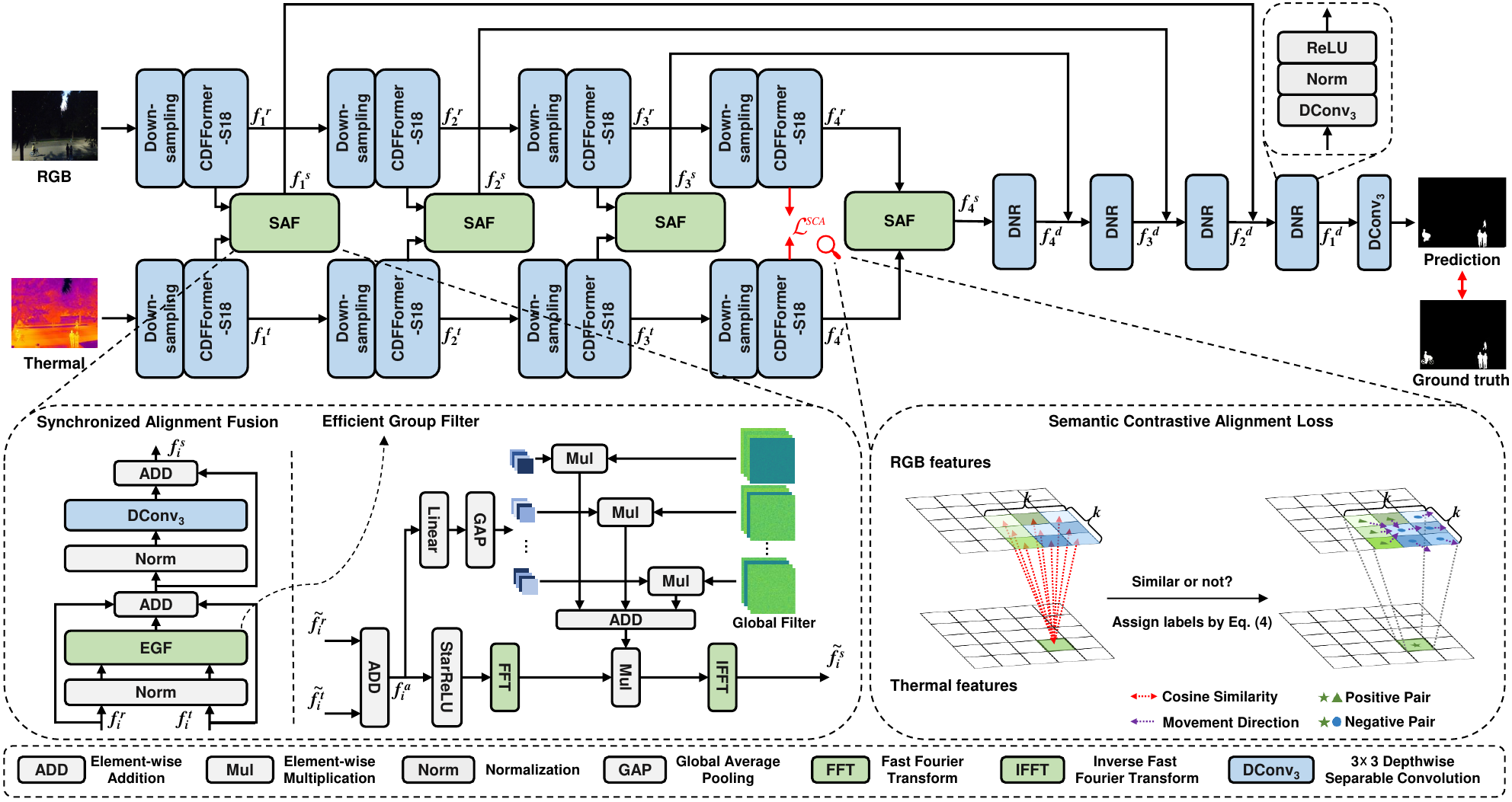}
	\caption{The overall framework of the proposed AlignSal. 
 The RGB and thermal images are first fed to a dual-stream encoder, 
 which extracts initial bimodal features $\left\{ {f_i^r} \right\}_{i = 1}^{4}$ and $\left\{ {f_i^t} \right\}_{i = 1}^{4}$. 
 During training, the semantic contrastive alignment loss (SCAL) facilitates the exchange of information between the RGB semantic feature $f_4^r$ and the thermal semantic feature $f_4^t$, 
 essentially achieving bimodal alignment. 
 The bimodal features are then aligned and fused at the pixel level by the synchronized alignment fusion (SAF). 
 Finally, a simple decoder integrates the fused features $\left\{ {f_i^s} \right\}_{i = 1}^{4}$ from high to low levels, 
 generating the decoder features $\left\{ {f_i^d} \right\}_{i = 1}^{4}$ and the final prediction $\mathcal{S}$.}
\label{fig:frame}
\end{figure*}

\section{Methods}
\subsection{Overview} 
As shown in Fig. \ref{fig:frame}, the proposed AlignSal consists of four main components: the dual-stream encoder, the semantic contrastive alignment loss (SCAL), the synchronized alignment fusion (SAF), and the decoder. 
First, considering that the fast Fourier transform (FFT) offers low complexity while capturing global dependencies, we use the lightweight FFT-based CDFFormer-S18 \cite{tatsunami2024fft} as the encoder. Leveraging the lower inductive bias introduced by FFT \cite{rao2021global,guibas2021adaptive,tatsunami2024fft,10127616}, 
we employ a Siamese architecture \cite{daudt2018fully,10127616} for the encoder to extract four levels of RGB features $\left\{ {f_i^r} \right\}_{i = 1}^{4}$ and thermal features $\left\{ {f_i^t} \right\}_{i = 1}^{4}$, further reducing complexity. 
During the training phase, the high-level bimodal features $f_{4}^r$ and $f_{4}^t$, rich in semantic information, are aligned by SCAL to achieve a consistent semantic distribution. 
Simultaneously, these bimodal features are fed into the corresponding SAF in pairs to achieve pixel-level alignment while capturing inter-modal complementary information. 
Finally, the outputs from the SAFs $\left\{ {f_i^s} \right\}_{i = 1}^{4}$ are passed through a simple decoder to progressively generate up-sampling features $\left\{ {f_i^d} \right\}_{i = 1}^{4}$ and the saliency map $\mathcal{S}$.

\subsection{Semantic contrastive alignment loss (SCAL)} \label{Sec:loss}
In AAV-based unaligned bimodal images, differences in resolution and fields of view lead to variations in the scale and position of salient objects in the spatial dimension. 
It is important that both modalities maintain consistent spatial position representation \cite{10315195}.
To address the challenge of aligning heterogeneous modalities, we note that contrastive learning \cite{radford2021learning,yang2021taco,he2023align,jiang2024t,song2024contrastalign} has recently demonstrated strong performance. Contrastive alignment \cite{jiang2024t,song2024contrastalign} is essentially a process of mutual refinement between modalities, where each modality benefits from knowledge exchange to enhance scene comprehension and achieve alignment.
In addition, the high-level features that contain rich semantic information can directly affect the model's understanding of the scene and the reconstruction of detailed features. 
With these factors in mind and without adding inference costs, we propose SCAL to achieve semantic-level spatial positional alignment of the two modalities and enhance single-modality representation through contrastive communication.

As illustrated in the diagram of SCAL in Fig. \ref{fig:frame}, since the receptive field of the thermal modality is small, we align each element in $f_{4}^t \in\mathbb{R}^{B \times C \times H \times W}$ as a base with the corresponding element in $f_{4}^r \in\mathbb{R}^{B \times C \times H \times W}$, where $B$, $C$, $H$, and $W$ represent batch size, channels, height, and width of features, respectively. First, we compress the number of channels of $f_{4}^r$ and $f_{4}^t$ to $1$ using a $3 \times 3$ depthwise separable convolution \cite{chollet2017xception} layer $DConv_{3}(\cdot)$ and a normalization layer $Norm(\cdot)$: 
\begin{equation}
\begin{aligned}
\tilde{f^r} = Norm(DConv_{3}(f_{4}^r)), %\thinspace \tilde{f^t} = Norm(DConv_{3}(f_{4}^{t})).
\end{aligned}
\end{equation}
\begin{equation}
\begin{aligned}
\tilde{f^t} = Norm(DConv_{3}(f_{4}^{t})).
\end{aligned}
\end{equation}
With the same resolution input, we observe that the region in RGB features corresponding to thermal features shifts and slightly shrinks in spatial position. Thus, given an element $a \in \mathbb{R}^{B \times C}$ with coordinate $(i,j)$ in $\tilde{f^t}$, we extract a patch $p \in \mathbb{R}^{B \times C \times k^{2}}$ using a window of size $k \times k$ centered on the corresponding position element in $\tilde{f^r}$. This strategy allows us to search for the element corresponding to $a$ within $p$, avoiding the need to search the entire $\tilde{f^r}$ that may significantly increase the computational burden during training.

Then, we calculate the similarity $s_{u}$ between $a$ and element $p_{u}$ in $p$ by the cosine similarity $Cosine(\cdot)$: 
\begin{equation}
\begin{aligned}
s_{u} = Cosine(a, p_{u}) = \frac{a \cdot p_{u}}{\|a\|_{2} \|p_{u}\|_{2}},
\end{aligned}
\end{equation}
where $\cdot$ is dot-product and $u \in \{1, 2, \ldots, k^{2}\}$ is the element index. Furthermore, we employ a threshold $t$ to classify the positive and negative samples in this $k^{2}$ region:
\begin{equation}
\begin{aligned}
\label{Eq:t}
l_{u} = 
\begin{cases} 
1 & \text{if } s_{u} > t \\ 
0 & \text{otherwise} 
\end{cases},
\end{aligned}
\end{equation}
where $l_{u}$ represents the label of $p_{u}$. Specifically, it is classified as a positive sample when $s_{u}$ is larger than $t$; otherwise, it is classified as negative. To bring positive samples closer to $a$ and push negative samples farther away, we employ InfoNCE \cite{oord2018representation} to measure the distance between samples:
\begin{equation}
\begin{aligned}
\mathcal{L}^{SCA} = - \frac{1}{HW} \sum_{i=1}^{H} \sum_{j=1}^{W} \sum_{u=1}^{k^{2}} l_{u} \log\left( \frac{\exp(s_{u})}{\sum_{v=1}^{k^{2}} \exp\left(s_{v}\right)} \right),
\end{aligned}
\end{equation}
where $v \in \{1, 2, \ldots, k^{2}\}$ is the element index. 
As training proceeds, the spatial representation of the RGB semantic features gradually converges with that of the thermal semantic features in the feature space. 
Benefiting from the mutual distillation characteristic of contrastive learning, the thermal semantic features are enhanced at the same time, with clearer salient regions.

\subsection{Synchronized Alignment Fusion (SAF)} \label{Sec:SAF}
Although bimodal features are aligned at deep layers, they still remain slightly offset at high-resolution shallow layers. In addition, the distribution of features varies significantly between modalities due to their distinct physical properties. Effectively fusing these complementary features is essential for achieving BSOD \cite{10315195,9926193,10003255,cong2023point,10042233,10127616}. 
To address these challenges, we propose SAF, which uses an FFT-based multiple-filtering strategy to fuse bimodal complementary features and align them in both spatial and channel dimensions.

As shown in Fig. \ref{fig:frame}, SAF consists of normalization layers, residual connections, a feed-forward network, and a token mixer.
To capture the local relevance, we equip a $DConv_{3}(\cdot)$ in the feed-forward network \cite{10015667}. For the token mixer, we introduce the efficient group filter (EGF). 
In EGF, we first integrate the normalized bimodal features $\tilde{f_{i}^r}$ and $\tilde{f_{i}^t}$ into the same feature $f_{i}^{a}$,  
reducing the computational cost for feature alignment and fusion. 
This process is denoted as:
\begin{equation}
\begin{aligned}
f_{i}^{a} = \tilde{f_{i}^r} + \tilde{f_{i}^t} \in \mathbb{R}^{B \times C \times H \times W},
\end{aligned}
\end{equation}
where $+$ is element-wise addition. 
After applying a StarReLU \cite{yu2022metaformer} layer $SR(\cdot)$, we perform an FFT on the integrated feature $f_{i}^{a}$, then multiply it by the group filter $\mathcal{G}(\cdot)$, where each channel is linearly coupled. 
The group channel weights in $\mathcal{G}(\cdot)$ are determined by applying global average pooling $GAP(\cdot)$ to $f_{i}^{a}$. The use of channel-weighted $\mathcal{G}(\cdot)$ enhances the sensitivity of capturing salient details and spatial shifts in the spatial dimension, thereby facilitating the fusion and alignment of bimodal features. 
Finally, the output $\tilde{f_i^s}$ of EGF is obtained by inverse FFT (IFFT). The whole process is expressed as:
\begin{equation}
\begin{aligned}
\tilde{f_i^s} = IFFT(\mathcal{G}(f_{i}^{a}) \times FFT(SR(f_{i}^{a}))),
\end{aligned}
\end{equation}
where $\times$ is element-wise multiplication. The $N$ global filters \cite{rao2021global} that make up $\mathcal{G}(\cdot)$ are denoted as $\{G_{1}, G_{2}, \ldots, G_{N}\}$. The channel-weighted $\mathcal{G}(\cdot) \in \mathbb{C}^{B \times C \times H \times (\frac{W}{2}+1)}$ is defined as:
\begin{equation}
\begin{aligned}
    \mathcal{G}(f_{i}^{a})_{:,c,:,:}=\sum_{n=1}^{N} \left(\dfrac{e^{r_{(c-1)N+n}}}{\sum_{m=1}^{N{e^{r_{(c-1)N+m}}}}}\right)G_n,
\end{aligned}
\end{equation}
where $c$ is the channel index, $n$ and $m$ are the filter indexes, and 
\begin{equation}
\begin{aligned}
(r_{1}, \ldots, r_{NC})^{\top} = GAP(Linear(f_{i}^{a})).
\end{aligned}
\end{equation}
In this paper, we set $N = 4$ to prevent over-computation.

\subsection{Decoder} \label{Sec:decoder}
In encoder-decoder structured networks, the distribution of features varies across levels. 
Low-level features capture basic patterns, such as edges and textures, while high-level features represent more abstract concepts and semantic information \cite{gao2021unified,wang2021cgfnet}. 
To utilize these diverse representations, we introduce a simple decoder that fits these features gradually from high to low levels. 
Each decoder layer consists of a $DConv_{3}(\cdot)$, a $Norm(\cdot)$, and ReLU function, denoted as $DNR(\cdot)$. The whole process is represented as:
\begin{equation}
\begin{aligned}
{f_i^d} = \left\{ {\begin{array}{*{20}{c}}
{DNR\left( {{f_{i}^s}} \right),i = 4}\\
{DNR\left( {Concat({f_{i}^s},U{p_{2}}({f_{i + 1}^d}))} \right),i = 1,2,3}
\end{array}} \right.
\end{aligned}
\end{equation}
where $Concat(\cdot)$ denotes the concatenation operation and $Up_2(\cdot)$ refers to up-sampling by a factor of 2. Finally, the saliency map $\mathcal{S}$ can be obtained by a $DConv_{3}(\cdot)$:
\begin{equation}
\begin{aligned}
\mathcal{S} = DConv_{3}(f_1^d).
\end{aligned}
\end{equation}

For the loss function, we employ the commonly used BCE \cite{de2005tutorial} and IoU \cite{mattyus2017deeproadmapper} losses to supervise the learning of $\mathcal{S}$, and employ SCAL to supervise the alignment of $f_4^r$ and $f_4^t$, which can be mathematically described as:
\begin{equation}
\begin{aligned}
\mathcal{L} = \beta_1\mathcal{L}^{BCE}(\mathcal{S}, \mathcal{G}) + \beta_2\mathcal{L}^{IoU}(\mathcal{S}, \mathcal{G}) + \mathcal{L}^{SCA}(f_4^r, f_4^t),
\end{aligned}
\end{equation}
where $\mathcal{G}$ stands for ground truth. $\left\{ {{\beta_i}} \right\}_{i = 1}^2$ represents weight parameters of different losses, and we set them to 1 in this paper.

\section{Experiments}
\subsection{Datasets}
To comprehensively validate our proposed AlignSal, we first conducted experiments on the AAV-based unaligned bimodal salient object detection (BSOD) task. To further assess its generalization, we extended our evaluation to three additional bimodal tasks: weakly aligned BSOD, aligned BSOD, and remote sensing change detection (RSCD).

\emph{1) AAV-based Unaligned BSOD:} We conducted extensive experiments on the AAV RGB-T 2400 dataset \cite{10315195}, which contains 2400 instantly captured unaligned RGB-T image pairs falling at different times of the day and different seasons of the year. Each image pair has a corresponding pixel-wise annotation. Following \cite{10315195}, all BSOD models were trained on the AAV RGB-T 2400 training set and tested on the testing set.

\emph{2) Weakly Aligned BSOD:} We employed unaligned-VT821, unaligned-VT1000, and unaligned-VT5000 datasets \cite{tu2022weakly} for the experiments. These three datasets contain 6821 RGB-T image pairs and were obtained by performing random affine transformations from the three corresponding aligned datasets VT821 \cite{wang2018rgb}, VT1000 \cite{tu2019rgb}, VT5000 \cite{9767629}. 
Following \cite{10315195,tu2022weakly}, all BSOD models were trained on the unaligned-VT5000 training set and tested on the remaining datasets.

\emph{3) Aligned BSOD:} We utilized VT821 \cite{wang2018rgb}, VT1000 \cite{tu2019rgb}, and VT5000 \cite{9767629} datasets, which collectively contain 6821 manually aligned RGB-T image pairs in the natural scene. Following \cite{10003255,10127616}, all BSOD models were trained on the VT5000 training set and tested on the remaining datasets.

\emph{4) RSCD:} To assess the generalization of AlignSal in high-altitude scenarios, we conducted experiments on the LEVIR-CD+ dataset \cite{chen2020spatial}, which comprises 985 remote sensing image pairs documenting significant building changes. Each image has an ultra-high resolution of 1024 $\times$ 1024 pixels at 0.5 meters per pixel. All RSCD models were trained on the dataset’s training set and evaluated on its testing set.

\subsection{Evaluation Metrics}
To fairly evaluate model performance for BSOD tasks, we used six widely recognized evaluation metrics. E-measure \cite{ijcai2018p97} ($Em$) assesses the joint contribution of pixel-level accuracy and image-level matching. S-measure \cite{fan2017structure} ($Sm$) combines region-aware and object-aware evaluation to quantify the structural similarity. F-measure \cite{5206596} (${F_{\beta}}$) balances precision and recall to provide a harmonic mean assessment of accuracy. While weighted F-measure \cite{6909433} (${wF_{\beta}}$) adjusts precision and recall by pixel importance for a refined saliency evaluation. Mean absolute error \cite{6247743} ($\mathcal{M}$) calculates the average pixel-wise absolute difference between prediction and ground truth. We also used Precision-recall (PR) curve to show the trade-off between precision and recall across various thresholds.

When evaluating model performance for the RSCD task, we employed five additional metrics \cite{chen2024changemamba}, including recall (Rec.), precision (Pre.), overall accuracy (OA), F1 score (F1), and IoU \cite{mattyus2017deeproadmapper}.

To evaluate model complexity, we further adopted three additional metrics: the number of parameters (Params), floating point operations (FLOPs), and frame per second (FPS).

\begin{table*}[!htp]
  \centering
  \fontsize{8}{10}\selectfont
  \renewcommand{\arraystretch}{1.0}
  \renewcommand{\tabcolsep}{1.8mm}
  \scriptsize
  %\captionsetup{labelformat=empty}
  \caption{Quantitative Comparison of E-measure ($Em$), S-measure ($Sm$), F-measure (${F_{\beta}}$), Weighted F-measure (${wF_{\beta}}$), and Mean Absolute Error ($\mathcal{M}$) Between Our AlignSal and Sixteen State-of-the-Art BSOD Models on the AAV RGB-T 2400 Dataset \cite{10315195}. Type A, U, and UU Stand for Aligned, Unaligned, and AAV-based Unaligned BSOD Models, Respectively. The Best Results are Labeled \textcolor{red}{\textbf{Red}} and the Second Best Results are Labeled \textcolor{blue}{Blue}. We Tested the FPS of Different Models on an NVIDIA RTX4060 Laptop GPU.}
\label{tab:comp_AAV}
  \scalebox{1}{
  \begin{tabular}{>{\centering\arraybackslash}p{2cm}|c|c|ccccc|ccc}
  \hline\toprule
   %\multirow{2}{*}{\centering Model} &  & \multicolumn{3}{c|}{\centering} & \multicolumn{5}{c}{\centering AAV RGB-T 2400}\\
   Model & Type &Backbone &$Em\uparrow$ & $Sm\uparrow$ &$wF_{\beta} \uparrow$ & $F_{\beta}\uparrow$ & $\mathcal{M}\downarrow$  & Params $\downarrow$ & FLOPs $\downarrow$ & FPS $\uparrow$\\
\midrule
    MIDD$_{21}$ \cite{9454273} & A & VGG-16 & 0.8333 & 0.8524 & 0.7168 & 0.6383 & 0.0114 & 52.4M & 216.7G & 86.3\\
    CSRNet$_{22}$ \cite{9505635} & A & ESPNetV2 & 0.8957 & 0.8015 & 0.6741 & 0.6943 & 0.0157 & 1.0M & 6.3G & 40.2\\	
    MoADNet$_{22}$ \cite{jin2022moadnet} & A & MobileNetV3 & 0.4627 & 0.4983 & 0.0591 & 0.0985 & 0.0370 & 5.0M & 1.3G & 44.0\\	
    OSRNet$_{22}$ \cite{9803225} & A & VGG-16 & 0.7731 & 0.7301 & 0.5027 & 0.4899 & 0.0290 & 42.4M & 15.6G & 95.6\\	
    SwinNet$_{22}$ \cite{9611276} & A & Swin-B & 0.8982 & 0.8756 & 0.8198 & 0.7160 & 0.0074 & 198.7M & 124.3G & 25.8 \\	
    LSNet$_{23}$ \cite{10042233} & A & MobileNetV2 & 0.8927 & 0.6947 & 0.3173 & 0.6796 & 0.0411 & 4.6M & 1.2G & 88.4 \\	
    TNet$_{23}$ \cite{9926193} & A &  ResNet-50 & 0.7972 & 0.6864 & 0.4452 & 0.4619 & 0.0266 & 87.0M & 39.7G & 59.0 \\
    HRTNet$_{23}$ \cite{9869666} & A & HRFormer & 0.9165 & 0.8644 & 0.7645 & 0.7115 & 0.0097 & 58.9M & 17.3G & 11.5\\
    MGAI$_{23}$ \cite{10003255} & A & Res2Net-50 & 0.8392 & 0.8296 & 0.6797 & 0.6083 & 0.0125 & 86.8M & 94.2G & 4.7 \\
    CATNet$_{23}$ \cite{sun2023catnet} & A & Swin-B & 0.7420 & 0.8155 & 0.6331 & 0.5189 & 0.0182 & 262.6M & 341.8G & 19.7\\
    PICRNet$_{23}$ \cite{cong2023point} & A & Swin-T & 0.8609 & 0.7685 & 0.6129 & 0.6083 & 0.0191 & 112.0M & 27.1G & 30.3\\
    WaveNet$_{23}$ \cite{10127616} & A & Wave-MLP & 0.9196 & 0.8929 & 0.8075 & 0.7294 & 0.0084 & 80.7M & 64.0G & 5.3\\
    CAVER$_{23}$ \cite{10015667} & A & ResNet-101 & 0.9551 & 0.8938 & 0.8218 & 0.7864 & 0.0066 & 93.8M & 63.9G & 16.2\\
    LAFB$_{24}$ \cite{wang2024learning} & A & Res2Net-50 & 0.9633 & \textcolor{blue}{0.9040} & 0.8471 & 0.8161 & 0.0068 & 453.0M & 139.7G & 29.3\\
\midrule
    DCNet$_{22}$ \cite{tu2022weakly} & U & VGG-16 & 0.9276 & 0.8292 & 0.7500 & 0.7997 & 0.0090 & 24.1M & 207.2G & 14.5\\	
    MROS$_{23}$ \cite{10315195}  & UU & Res2Net-50 &\textcolor{red}{\textbf{0.9747}} & 0.8963 & \textcolor{blue}{0.8506} & \textcolor{red}{\textbf{0.8494}} & \textcolor{blue}{0.0063} & 90.4M & 45.3G & 12.2 \\
\midrule
    AlignSal (Ours) & UU & CDFFormer-S18 & \textcolor{blue}{0.9654} & \textcolor{red}{\textbf{0.9085}} & \textcolor{red}{\textbf{0.8599}} & \textcolor{blue}{0.8277} & \textcolor{red}{\textbf{0.0062}} & 27.1M & 22.9G & 30.8 \\
    \bottomrule
    \hline
  \end{tabular}}
\end{table*}

\begin{table*}[!htp]
  \centering
  \fontsize{8}{10}\selectfont
  \renewcommand{\arraystretch}{1.1}
  \renewcommand{\tabcolsep}{0.5mm}
  \scriptsize
  %\captionsetup{labelformat=empty}
  \caption{Quantitative Comparison Between AlignSal and Existing State-of-the-Art BSOD Models in Various Challenging Scenes. The Best Results are Labeled \textcolor{red}{\textbf{Red}} and the Second Best Results are Labeled \textcolor{blue}{Blue}.}
\label{tab:challenge}
  \scalebox{1}{
  \begin{tabular}{>{\centering\arraybackslash}p{1.8cm}|ccccc|ccccc|ccccc|ccccc}
  \hline\toprule
   \multirow{2}{*}{\centering Model} & \multicolumn{5}{c|}{\centering Object blurring} & \multicolumn{5}{c|}{\centering Illumination changing} & \multicolumn{5}{c|}{\centering Object changing} & \multicolumn{5}{c}{\centering Weather changing}\\
   & $Em\uparrow$ & $Sm\uparrow$ & $wF_{\beta} \uparrow$ & $F_{\beta}\uparrow$ & $\mathcal{M}\downarrow$  $\uparrow$ &$Em\uparrow$ & $Sm\uparrow$ &$wF_{\beta} \uparrow$ & $F_{\beta}\uparrow$ & $\mathcal{M}\downarrow$ & $Em\uparrow$ & $Sm\uparrow$ & $wF_{\beta} \uparrow$ & $F_{\beta}\uparrow$ & $\mathcal{M}\downarrow$ & $Em\uparrow$ & $Sm\uparrow$ & $wF_{\beta} \uparrow$ & $F_{\beta}\uparrow$ & $\mathcal{M}\downarrow$ \\
\midrule
    MIDD$_{21}$ \cite{9454273} & 0.8669 & 0.8455 & 0.7260 & 0.6754 & 0.0116 & 0.8415 & 0.8543 & 0.7270 & 0.6467 & 0.0110 & 0.8215 & 0.8332 & 0.6934 & 0.6228 & 0.0132 & 0.8861 & 0.9149 & 0.8152 & 0.7014 & 0.0045\\
    CSRNet$_{22}$ \cite{9505635} & 0.8971 & 0.7849 & 0.6690 & 0.7175 & 0.0186 & 0.8919 & 0.7924 & 0.6655 & 0.6927 & 0.0154 & 0.8793 & 0.7750 & 0.6355 & 0.6654 & 0.0185 & 0.9542 & 0.9056 & 0.8415 & 0.8262 & 0.0050\\	
    MoADNet$_{22}$ \cite{jin2022moadnet} & 0.4659 & 0.5057 & 0.0806 & 0.1354 & 0.0389 & 0.4609 & 0.4985 & 0.0563 & 0.0974 & 0.0376 & 0.4708 & 0.4993 & 0.0649 & 0.1055 & 0.0395 & 0.4389 & 0.4398 & 0.0192 & 0.0422 & 0.0266\\	
    OSRNet$_{22}$ \cite{9803225} & 0.8106 & 0.7414 & 0.5356 & 0.5378 & 0.0279 & 0.7777 & 0.7327 & 0.5150 & 0.5064 & 0.0284 & 0.7599 & 0.7130 & 0.4810 & 0.4744 & 0.0325 & 0.8642 & 0.7851 & 0.6052 & 0.5759 & 0.0103\\	
    SwinNet$_{22}$ \cite{9611276} & 0.9293 & 0.9080 & 0.8553 & 0.7687 & 0.0059 & 0.8980 & 0.8993 & 0.8347 & 0.7280 & 0.0074 & 0.8860 & 0.8801 & 0.8030 & 0.6966 & 0.0088 & 0.9344 & 0.9473 & 0.8956 & 0.7778 & 0.0025\\	
    LSNet$_{23}$ \cite{10042233} & 0.9111 & 0.7201 & 0.3632 & 0.7127 & 0.0378 & 0.8973 & 0.6932 & 0.3222 & 0.6897 & 0.0407 & 0.8828 & 0.6790 & 0.3124 & 0.6598 & 0.0450 & 0.9315 & 0.7268 & 0.3172 & 0.7613 & 0.0267\\	
    TNet$_{23}$ \cite{9926193} & 0.7998 & 0.6550 & 0.3913 & 0.4353 & 0.0310 & 0.7966 & 0.6969 & 0.4675 & 0.4853 & 0.0255 & 0.7826 & 0.6767 & 0.4338 &	0.4525 & 0.0296 & 0.8727 & 0.7734 & 0.5884 & 0.5964 & 0.0112\\
    HRTNet$_{23}$ \cite{9869666} & 0.9289 & 0.8711 & 0.7825 & 0.7423 & 0.0091 & 0.9162 & 0.8668 & 0.7719 & 0.7183 & 0.0094 & 0.9069 & 0.8490 & 0.7420 & 0.6914 & 0.0112 & 0.9427 & 0.9192 & 0.8531 & 0.7803 & 0.0035\\
    MGAI$_{23}$ \cite{10003255} & 0.8722 & 0.8406 & 0.7121 & 0.6559 & 0.0121 & 0.8374 & 0.8300 & 0.6863 & 0.6127 & 0.0124 & 0.8245 & 0.8091 & 0.6537 & 0.5851 & 0.0145 & 0.8691 & 0.8827 & 0.7575 & 0.6540 & 0.0059\\
    CATNet$_{23}$ \cite{sun2023catnet} & 0.7734 & 0.8331 & 0.6627 & 0.5645 & 0.0183 & 0.7511 & 0.8227 & 0.6520 & 0.5393 & 0.0166 & 0.7281 & 0.7953 & 0.6079 & 0.5021 & 0.0207 & 0.8031 & 0.8946 & 0.7514 & 0.5975 & 0.0054\\
    PICRNet$_{23}$ \cite{cong2023point} & 0.8678 & 0.7632 & 0.6067 & 0.6233 & 0.0239 & 0.8600 & 0.7671 & 0.6175 & 0.6101 & 0.0185 & 0.8469 & 0.7461 & 0.5842 & 0.5830 & 0.0219 & 0.9206 & 0.8566 & 0.7510 & 0.7240 & 0.0061\\
    WaveNet$_{23}$ \cite{10127616} & 0.9462 & 0.9064 & 0.8511 & 0.7782 & 0.0063 & 0.9246 & 0.9009 & 0.8253 & 0.7433 & 0.0083 & 0.9100 & 0.8790 & 0.7891 & 0.7115 & 0.0099 & 0.9497 & 0.9486 & 0.8894 & 0.7929 & 0.0027\\
    CAVER$_{23}$ \cite{10015667} & 0.9657 & 0.9007 & 0.8416 & 0.8076 & 0.0068 & 0.9531 & 0.8953 & 0.8252 & 0.7862 & 0.0066 & 0.9509 & 0.8829 & 0.8093 & 0.7748 & \textcolor{blue}{0.0075} & 0.9642 & 0.9333 & 0.8762 & 0.8236 & 0.0030\\
    LAFB$_{24}$ \cite{wang2024learning} & 0.9769 & \textcolor{blue}{0.9142} & \textcolor{blue}{0.8759} & 0.8531 & \textcolor{blue}{0.0057} & 0.9673 & \textcolor{blue}{0.9115} & \textcolor{blue}{0.8650} & 0.8332 & 0.0067 & 0.9576 & \textcolor{blue}{0.8914} & 0.8297 & 0.7990 & 0.0081 & 0.9819 & \textcolor{blue}{0.9511} & \textcolor{blue}{0.9199} & 0.8842 & \textcolor{blue}{0.0020}\\
\midrule
    DCNet$_{22}$ \cite{tu2022weakly} & 0.9529 & 0.8369 & 0.7748 & 0.8229 & 0.0090 & 0.9424 & 0.8314 & 0.7639 & 0.8134 & 0.0089 & 0.9185 & 0.8120 & 0.7285 & 0.7839 & 0.0105 & 0.9740 & 0.8886 & 0.8490 & 0.8813 & 0.0038\\	
    MROS$_{23}$ \cite{10315195} &\textcolor{blue}{0.9776} & 0.9002 & 0.8582 & \textcolor{blue}{0.8577} & 0.0065 &\textcolor{red}{\textbf{0.9733}} & 0.8951 & 0.8514 & \textcolor{red}{\textbf{0.8494}} & \textcolor{blue}{0.0064} & \textcolor{red}{\textbf{0.9706}} & 0.8841 & \textcolor{blue}{0.8365} & \textcolor{red}{\textbf{0.8383}} & \textcolor{red}{\textbf{0.0074}} & \textcolor{red}{\textbf{0.9883}} & 0.9404 & 0.9121 & \textcolor{red}{\textbf{0.9009}} & 0.0022\\
\midrule
    AlignSal (Ours) & \textcolor{red}{\textbf{0.9777}} & \textcolor{red}{\textbf{0.9158}} & \textcolor{red}{\textbf{0.8886}} & \textcolor{red}{\textbf{0.8580}} & \textcolor{red}{\textbf{0.0047}} & \textcolor{blue}{0.9700} & \textcolor{red}{\textbf{0.9178}} & \textcolor{red}{\textbf{0.8790}} & \textcolor{blue}{0.8431} & \textcolor{red}{\textbf{0.0060}} & \textcolor{blue}{0.9591} & \textcolor{red}{\textbf{0.8951}} & \textcolor{red}{\textbf{0.8424}} & \textcolor{blue}{0.8085} & \textcolor{red}{\textbf{0.0074}} & \textcolor{blue}{0.9860} & \textcolor{red}{\textbf{0.9536}} & \textcolor{red}{\textbf{0.9281}} & \textcolor{blue}{0.8954} & \textcolor{red}{\textbf{0.0019}} \\
    \bottomrule
    \hline
  \end{tabular}}
\end{table*}

\subsection{Implementation Details}
We trained our model on an NVIDIA Tesla P100 GPU, using the Adam optimizer \cite{kingma2014adam} with a batch size of 8. Our model utilized the CDFFormer-S18 \cite{tatsunami2024fft} as the encoder, which is pre-trained on ImageNet. The initial learning rate was set to $3\times10^{-5}$ and decreased by a factor of 10 every 100 epochs, over a total of 300 epochs. The input resolution was set to $384\times384$. 
\begin{table}[!htp]
\centering
  \fontsize{8}{10}\selectfont
  \renewcommand{\arraystretch}{1}
  \renewcommand{\tabcolsep}{0.8mm}
  \scriptsize
  \caption{Selection of Window Size $k$ and Similarity Threshold $t$ in Semantic Contrastive Alignment Loss. The Best Results are Labeled \textcolor{red}{\textbf{Red}}.}
  \label{tab:scal_hyperparam}
  \scalebox{1}{
  \begin{tabular}{c|ccccc}
  \hline\toprule
  Setting &$Em\uparrow$ & $Sm\uparrow$ &$wF_{\beta} \uparrow$ & $F_{\beta}\uparrow$ & $\mathcal{M}\downarrow$\\
\midrule
  $k=3$, $t=0.4$ & 0.9654 & 0.9085 & 0.8599 & \textcolor{red}{\textbf{0.8277}} & \textcolor{red}{\textbf{0.0062}}\\
  $k=5$, $t=0.4$ & 0.9652 & 0.9085 & \textcolor{red}{\textbf{0.8605}} & 0.8251 & 0.0063\\
\midrule
  $t=0.5$, $k=3$ & \textcolor{red}{\textbf{0.9662}} & 0.9059 & 0.8578 & 0.8271 & 0.0065\\
  $t=0.3$, $k=3$ & 0.9595 & \textcolor{red}{\textbf{0.9094}} & 0.8588 & 0.8137 & 0.0066\\
  \bottomrule
  \hline
  \end{tabular}}
\end{table}
Additionally, we applied various data augmentation techniques to mitigate overfitting during training, including random cropping, horizontal flipping, and multi-angle rotations.

For the hyperparameters in SCAL, 
namely the window size $k$ and the threshold $t$ in Eq. (\ref{Eq:t}), 
after detailed testing, 
we decided to set $k$ to 3 and $t$ to 0.4 in our experiments, 
unless specified otherwise. 
Representative values of $k$ and $t$ in the testing are shown in Table \ref{tab:scal_hyperparam}. 

\subsection{Model Comparison}
We compared our AlignSal with fourteen state-of-the-art aligned BSOD models, one state-of-the-art unaligned BSOD model, and one state-of-the-art AAV-based unaligned BSOD model. The aligned BSOD models include MIDD \cite{9454273}, CSRNet \cite{9505635}, MoADNet \cite{jin2022moadnet}, SwinNet \cite{9611276}, OSRNet \cite{9803225}, TNet \cite{9926193}, MGAI \cite{10003255}, CATNet \cite{sun2023catnet}, PICRNet \cite{cong2023point}, LSNet \cite{10042233}, HRTNet \cite{9869666}, CAVER \cite{10015667}, WaveNet \cite{10127616}, and LAFB \cite{wang2024learning}. The unaligned BSOD model is DCNet \cite{tu2022weakly}. The AAV-based unaligned BSOD model is MROS \cite{10315195}. All saliency maps were obtained from the authors or generated by the open-source code and were evaluated by the Saliency Evaluation Toolbox\footnote{https://github.com/jiwei0921/Saliency-Evaluation-Toolbox}.

{\textbf{Quantitative Evaluation.}} To fairly evaluate the performance of our AlignSal on the AAV RGB-T 2400 dataset \cite{10315195}, we showed the comparison of $Em$, $Sm$, $wF_{\beta}$, $F_{\beta}$, $\mathcal{M}$,  
\begin{figure*}[!htp]
	\centering \includegraphics[width=1\textwidth]{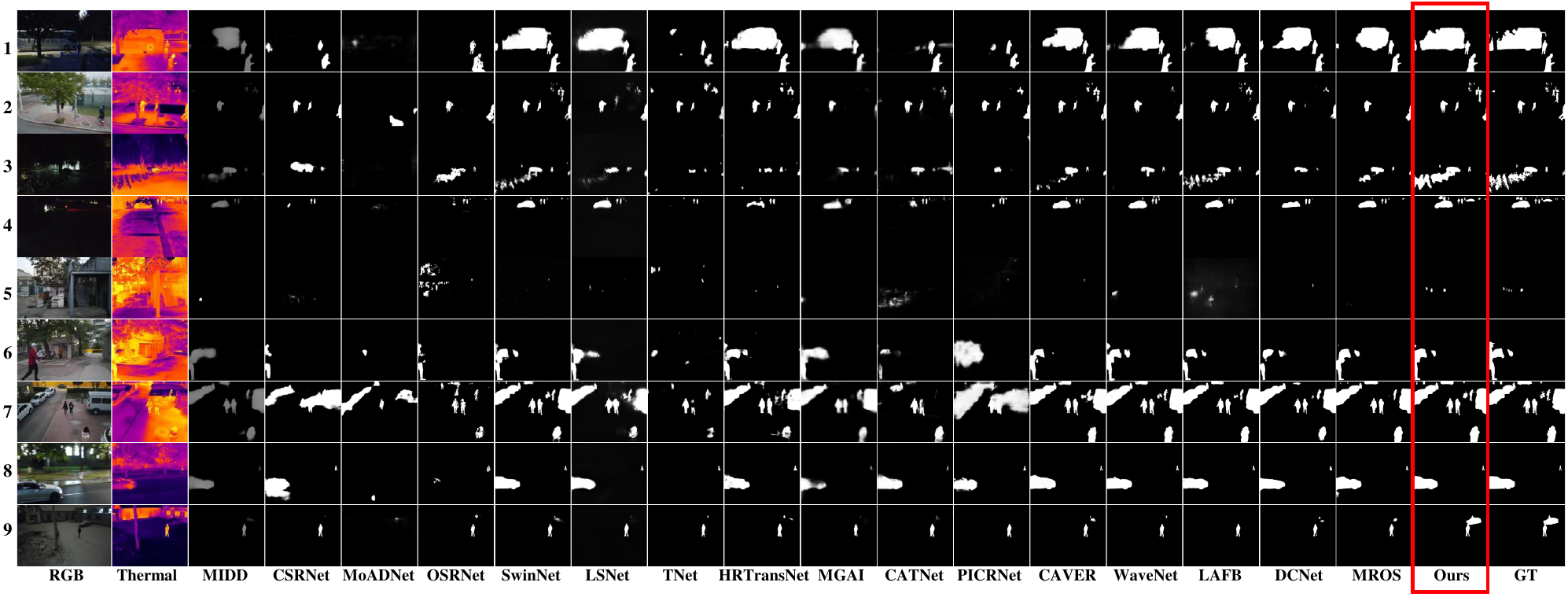}
	\caption{Visual comparison of our AlignSal with state-of-the-art BSOD models in several challenges, including fast AAV movement ($1^{st}$, $2^{nd}$, and $6^{th}$ rows) and fast object movement ($8^{th}$ row) in object blurring scenes, low illumination ($1^{st}$ and $8^{th}$ rows), street light exposure ($3^{rd}$ and $9^{th}$ rows), and extreme low illumination ($4^{th}$ row) in illumination changing scenes, small objects
    ($2^{nd}$, $4^{th}$, $5^{th}$, and $8^{th}$ rows), out-of-view ($1^{st}$, $2^{nd}$, $6^{th}$, and $8^{th}$ rows), multiple objects($1^{st}$-$4^{th}$, $7^{th}$-$9^{th}$ rows), scale variation ($8^{th}$ row), and center bias ($4^{th}$ and $6^{th}$ rows) in object changing scenes, rain ($8^{th}$ row) and snow ($9^{th}$ row) in weather changing scenes. GT represents ground truth.}
\label{fig:comp}
\end{figure*}
\begin{figure}[!htp]
	\centering \includegraphics[width=0.485\textwidth]{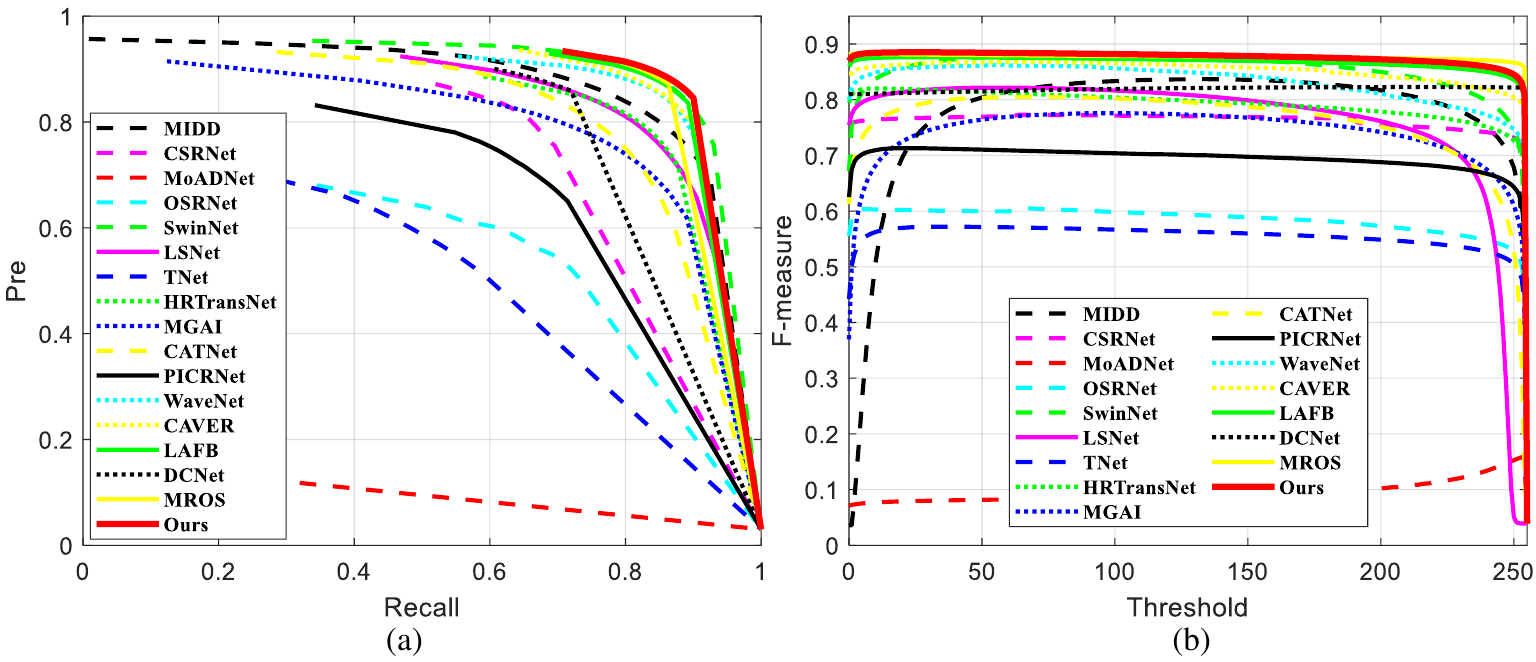}
	\caption{(a) Precision-recall (PR) and (b) F-measure-threshold (FT) curves of BSOD models on the AAV RGB-T 2400 dataset.}
\label{fig:curve}
\end{figure}
and three complexity metrics (\textit{i.e.}, Params, FLOPs, and FPS) of our AlignSal with the above BSOD models in Table \ref{tab:comp_AAV}. Notably, with real-time inference speed and lower complexity, our model achieved superior performance compared to the state-of-the-art BSOD models. 
Specifically, compared to the existing top-performing AAV-based unaligned model MROS \cite{10315195}, 
our model surpassed it in three out of five metrics (\textit{i.e.} an improvement of 1.4\%, 1.1\%, and 1.6\% in $Sm$, $wF_{\beta}$, and $\mathcal{M}$, respectively),
while being much lighter (\textit{i.e.} 70\% reduction in Params) and having much faster inference speed (\textit{i.e.} half the FLOPs and 2.5 times the FPS). 
Compared to the existing best aligned BSOD model LAFB \cite{wang2024learning}, 
our model outperformed it by 0.2\%, 0.5\%, 1.5\%, 1.4\%, and 8.8\% in $Em$, $Sm$, $wF_{\beta}$, $F_{\beta}$, and $\mathcal{M}$, respectively,
while having a 94.1\% and 83.6\% reduction of Params and FLOPs. In addition, we also provided the comparison of PR and F-measure-threshold (FT) curves in Fig. \ref{fig:curve}. We can observe that the PR curve of our model is closest to (1,1), and the area under the FT curve is the largest as well. 

To validate the robustness of our AlignSal, we conducted extensive experiments across various challenging scenes, including object blur, illumination changes, object variations (\textit{i.e.} small object, multiple objects, center bias, out-of-view, and scale variation), and weather changes. 
As shown in Table \ref{tab:challenge}, 
our model outperformed all the existing BSOD models in object blur scenes across all five metrics. 
In the other three challenging scenes, 
our AlignSal still beat other models in most metrics (\textit{i.e.} outperforming MROS \cite{10315195} by 1.7\%, 1.9\%, and 4.4\% in $Sm$, $wF_{\beta}$, and $\mathcal{M}$ on average, respectively).
% and $Em$ and $F_{\beta}$ averaging 0.6\% and 1.6\% lower. 

To summarize, the above experiments demonstrated that our AlignSal can effectively cope with challenging and extreme scenes. 
More importantly, AlignSal balances performance and complexity, achieving superior performance with real-time inference speed and low computational expense.

{\textbf{Qualitative Comparison.}} 
As shown in Fig. \ref{fig:comp}, 
we present a visual comparison of our AlignSal with existing state-of-the-art BSOD models across several challenging scenes, 
including fast AAV movement ($1^{st}$, $2^{nd}$, and $6^{th}$ rows) and fast object movement ($8^{th}$ row) in object blurring scenes; 
low illumination ($1^{st}$ and $8^{th}$ rows), 
street light exposure ($3^{rd}$ and $9^{th}$ rows), 
and extreme low illumination ($4^{th}$ row) in illumination changing scenes; 
small objects ($2^{nd}$, $4^{th}$, $5^{th}$, and $8^{th}$ rows), 
out-of-view ($1^{st}$, $2^{nd}$, $6^{th}$, and $8^{th}$ rows), 
multiple objects($1^{st}$-$4^{th}$, $7^{th}$-$9^{th}$ rows), 
scale variation ($8^{th}$ row), 
and center bias ($4^{th}$ and $6^{th}$ rows) in object changing scenes; 
rain ($8^{th}$ row) and snow ($9^{th}$ row) in weather changing scenes. 
As shown, a scene often contains multiple challenges and AlignSal can completely capture objects in these extreme scenes. 
This is attributed to our proposed SCAL and SAF and the way they cooperate within AlignSal. 
\begin{table}[!htp]
\centering
  \fontsize{8}{10}\selectfont
  \renewcommand{\arraystretch}{1}
  \renewcommand{\tabcolsep}{0.8mm}
  \scriptsize
  \caption{Ablation Studies of the Key Components in AlignSal on the AAV RGB-T 2400 Dataset. The Best Results are Labeled \textcolor{red}{\textbf{Red}}.}
  \label{tab:ablation}
  \scalebox{1}{
  \begin{tabular}{c|ccccc|cc}
  \hline\toprule
  Setting  &$Em\uparrow$ & $Sm\uparrow$ &$wF_{\beta} \uparrow$ & $F_{\beta}\uparrow$ & $\mathcal{M}\downarrow$ & Params $\downarrow$ & FLOPs $\downarrow$\\
\midrule
  w/ GFNet-H-S & 0.8932 & 0.8035 & 0.6613 & 0.6731 & 0.0156 & 28.7M & 27.1G\\
  w/ AFNO & 0.7743 & 0.7695 & 0.5429 & 0.5142 & 0.0204 & 23.2M & 22.3G\\
  w/ DFFormer-S18 & 0.9637 & \textcolor{red}{\textbf{0.9091}} & 0.8528 & 0.8144 & 0.0068 & 27.2M & 22.4G \\
  w/ ResNet-50 & 0.5489 & 0.5892 & 0.3307 & 0.3837 & 0.0402 & 23.6M & 24.5G \\
  w/ Swin-T & 0.7378 & 0.7219 & 0.5953 & 0.5847 & 0.0174 & 27.5M & 25.9G \\
\midrule
  w/o SCAL & 0.9583 & 0.8983 & 0.8207 & 0.7996 & 0.0071  & 27.1M & 22.9G\\
  w/o SAF & 0.9470 &  0.8835 & 0.8295 & 0.7860 & 0.0074 & 27.1M & 22.8G\\
\midrule
  Full model & \textcolor{red}{\textbf{0.9654}} & 0.9085 & \textcolor{red}{\textbf{0.8599}} & \textcolor{red}{\textbf{0.8277}} & \textcolor{red}{\textbf{0.0062}}  & 27.1M & 22.9G\\
  \bottomrule
  \hline
  \end{tabular}}
\end{table}
SCAL aligns and mutually refines bimodal features at the semantic level using a contrastive strategy, 
while SAF aligns the two modalities at the pixel level by using the powerful filtering mechanism to effectively explore the bimodal complementary information and capture fine details.

\subsection{Ablation Study} \label{sec.abl}
To assess the contribution of each key component, namely the backbone network, SCAL, and SAF, in our proposed AlignSal, 
we conducted ablation experiments on the AAV RGB-T 2400 dataset.

\emph{1) Effectiveness of backbone network:} Owing to its efficient and powerful feature-generating capability, we chose the FFT-based CDFFormer-S18 \cite{tatsunami2024fft} as the backbone for our AlignSal. To verify its effectiveness, we selected three other FFT-based backbones to replace CDFFormer-S18, including GFNet-H-S \cite{rao2021global}, AFNO \cite{guibas2021adaptive}, and DFFormer-S18 \cite{tatsunami2024fft}. 
While AFNO had the same network depth as CDFFormer-S18, 
the comparison in Table \ref{tab:ablation} showed that CDFFormer-S18 has significant advantages over other models. 

To further assess the compatibility of AlignSal with spatial domain-based backbone networks, we conducted experiments using 
ResNet-50 \cite{he2016deep} and Swin Transformer Tiny (Swin-T) \cite{liu2021swin} as backbone networks. 
Table \ref{tab:ablation} demonstrates that AlignSal can be effectively integrated with these spatial domain-based backbone networks. However, our findings also show that their performance is weaker than FFT-based backbone networks when computational resources are limited.

\begin{figure}[!htp]
	\centering \includegraphics[width=0.485\textwidth]{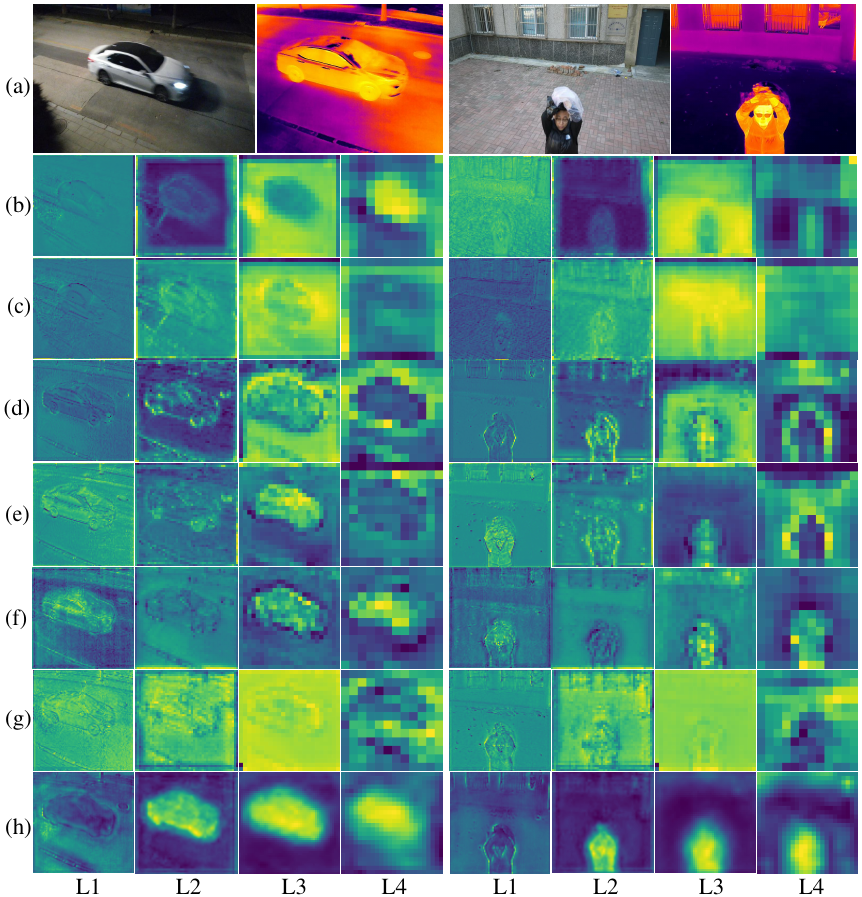}
	\caption{Visualization of the feature maps generated by models with and without SCAL and SAF. L1 to L4 represents the levels from low to high. (a) displays RGB and thermal images. (b) and (d) present the RGB and thermal feature maps, while (c) and (e) show the RGB and thermal feature maps generated by the model without SCAL. (f) shows the feature maps after SAF. (g) illustrates the fused feature maps from the model without SAF. (h) shows the feature maps after the decoder layers.}
\label{fig:attn}
\end{figure}
\begin{figure}[!htp]
	\centering \includegraphics[width=0.48\textwidth]{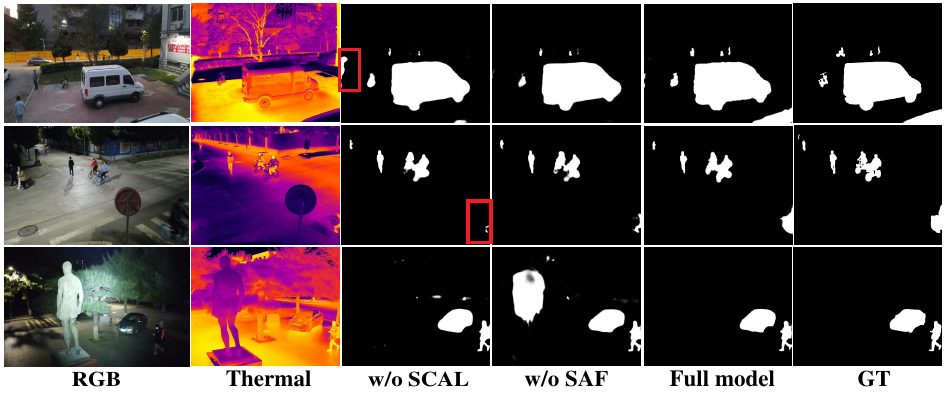}
	\caption{Visual comparison of the model with and without (w/o) the key components on the AAV RGB-T 2400 dataset. GT represents ground truth.}
\label{fig:ablation}
\end{figure}

\emph{2) Effectiveness of SCAL:} To assess the specific contribution of SCAL, we conducted ablation experiments from various perspectives. First, we removed SCAL and presented the results in Table \ref{tab:ablation}, Fig. \ref{fig:attn}, and Fig. \ref{fig:ablation}. 
The $Em$, $Sm$, $wF_{\beta}$, $F_{\beta}$, and $\mathcal{M}$ of the model without SCAL,
as shown in Table \ref{tab:ablation},
decreased significantly by 0.7\%, 1.1\%, 4.6\%, 3.4\%, and 14.5\%, respectively. 
Missing the semantic-level alignment resulted in inconsistencies in the salient regions highlighted by the two modalities, adversely affecting the final predictions. 
Specifically, comparing (b) and (c) in Fig. \ref{fig:attn}, 
the salient regions in the RGB feature maps of the model without SCAL were less distinct and exhibited noticeable drifts. 
In contrast, the model with SCAL not only resolved these issues, but also, as illustrated by comparing (d) and (e) in Fig. \ref{fig:attn}, enabled the thermal features to benefit from the RGB features, better distinguishing salient regions from the background. 
Additionally, SCAL encouraged the model to reuse low-level features more efficiently, improving feature representation and reducing the burden of subsequent pixel-level alignment. 
In the results of the model without SCAL shown in the $1^{st}$ row of Fig. \ref{fig:ablation}, the leftmost ground area (\textit{i.e.}, the area in red box) was incorrectly predicted as salient due to its highly distinctive features in the thermal modality. 
Similarly, the rightmost bike in the $2^{nd}$ row was misclassified as the background.

Second, to further validate the performance of SCAL in enhancing the model to handle unaligned data, 
we incorporated SCAL into the training phase of representative aligned BSOD models (\textit{i.e.}, MIDD \cite{9454273}, MoADNet \cite{jin2022moadnet}, LSNet \cite{10042233}, MGAI \cite{10003255}, and CATNet \cite{sun2023catnet}) on the AAV RGB-T 2400 dataset. 
In this experiment, we set the values of $k$ and $t$ of SCAL to 3 and 0.5, respectively, without fine-tuning. 
As seen in Table \ref{tab:state-of-the-art_w_scal}, the performance gains from SCAL were substantial. 
The $Em$, $Sm$, $wF_{\beta}$, $F_{\beta}$, and $\mathcal{M}$ improved by an average of 2.0\%, 2.1\%, 7.0\%, 4.7\%, and 16.2\%, respectively.

To summarize, the ablation experiments in this section demonstrated the high effectiveness of SCAL. 
It is a crucial component of our AlignSal and facilitates the transformation of aligned models into unaligned models,
\begin{table*}[!htp]
  \centering
  \fontsize{8}{10}\selectfont
  \renewcommand{\arraystretch}{1.1}
  \renewcommand{\tabcolsep}{0.8mm}
  \scriptsize
  %\captionsetup{labelformat=empty}
  \caption{Quantitative Comparison Between Our AlignSal and Eleven State-of-the-Art BSOD Models on the Unaligned-VT821, Unaligned-VT1000, and Unaligned-VT5000 Datasets \cite{tu2022weakly}. The Best Results are Labeled \textcolor{red}{\textbf{Red}} and the Second Best Results are Labeled \textcolor{blue}{Blue}. We Tested the FPS of Different Models on an NVIDIA RTX4060 Laptop GPU.}
\label{tab:generalization}
  \scalebox{1}{
  \begin{tabular}{>{\centering\arraybackslash}p{1.8cm}|ccccc|ccccc|ccccc|ccc}
  \hline\toprule
   \multirow{2}{*}{\centering Model} & \multicolumn{5}{c|}{\centering unaligned-VT821} & \multicolumn{5}{c|}{\centering unaligned-VT1000} & \multicolumn{5}{c|}{\centering unaligned-VT5000} & \multicolumn{3}{c}{\centering} \\
    &$Em\uparrow$ & $Sm\uparrow$ &$wF_{\beta} \uparrow$ & $F_{\beta}\uparrow$ & $\mathcal{M}\downarrow$ & $Em\uparrow$ & $Sm\uparrow$ & $wF_{\beta} \uparrow$ & $F_{\beta}\uparrow$ & $\mathcal{M}\downarrow$ & $Em\uparrow$ & $Sm\uparrow$ & $wF_{\beta} \uparrow$ & $F_{\beta}\uparrow$ & $\mathcal{M}\downarrow$ & Params $\downarrow$ & FLOPs $\downarrow$ & FPS $\uparrow$\\
\midrule
    MIDD$_{21}$ \cite{9454273} & 0.8806 & 0.8626 & 0.7473 & 0.7803 & 0.0512 & 0.9208 & 0.9062 & 0.8356 & 0.8616 & 0.0318 & 0.8850 & 0.8550 & 0.7429 & 0.7782 & 0.0489 & 52.4M & 216.7G & 86.3 \\
    MoADNet$_{22}$ \cite{jin2022moadnet} & 0.8060 & 0.7572 & 0.6016 & 0.6579 & 0.1071 & 0.8606 & 0.8199 & 0.7109 & 0.7696 & 0.0673 & 0.8266 & 0.7706 & 0.6096 & 0.6742 & 0.0795 & 5.0M & 1.3G & 44 \\	
    OSRNet$_{22}$ \cite{9803225} & 0.8300 & 0.8000 & 0.6917 & 0.7096 & 0.0655 & 0.9116 & 0.8928 & 0.8465 & 0.8557 & 0.0320 & 0.8780 & 0.8305 & 0.7414 & 0.7672 & 0.0498 & 42.4M & 15.6G & 95.6 \\	
    LSNet$_{23}$ \cite{10042233}& 0.8719 & 0.8496 & 0.7354 & 0.7633 & 0.0489 & 0.9096 & 0.9064 & 0.8396 & 0.8496 & 0.0312 & 0.8897 & 0.8564 & 0.7561 & 0.7762 & 0.0454  & 4.6M & 1.2G & 88.4 \\	
    TNet$_{23}$ \cite{9926193} & 0.8806 & 0.8626 & 0.7473 & 0.7803 & 0.0512 & 0.9208 & 0.9062 & 0.8356 &	0.8616 &	0.0318 & 0.8850 & 0.8550 & 0.7429 & 0.7782 & 0.0489 & 87.0M & 39.7G & 59 \\
    MGAI$_{23}$ \cite{10003255} & 0.8872 & 0.8599 & 0.7594 & 0.7762 & 0.0429 & 0.9264 & 0.9214 & 0.8652 & 0.8541 &	0.0268 & 0.8989 & 0.8645 & 0.7706 & 0.7865 & 0.0420 & 86.8M & 94.2G & 4.7 \\
    CATNet$_{23}$ \cite{sun2023catnet} & 0.8934 & 0.8596 & 0.7719 & 0.7899 & 0.0404 & 0.9289 & 0.9203 & 0.8855 & 0.8787 & 0.0224 & 0.9137 & 0.8705 & 0.8065 & 0.8176 & 0.0352 & 262.6M & 341.8G & 19.7 \\
    PICRNet$_{23}$ \cite{cong2023point} &0.8339 & 0.7917 & 0.6644 & 0.6918 & 0.0696 & 0.8545 & 0.8285 & 0.7374 & 0.7536 & 0.0670 & 0.8084 & 0.7610 & 0.6243 & 0.6517 & 0.0920 & 112.0M & 27.1G & 30.3 \\
    LAFB$_{24}$ \cite{wang2024learning}  & 0.8917 & 0.8588 & 0.7772 & 0.7959 & 0.0502 & \textcolor{blue}{0.9420} & \textcolor{blue}{0.9264} & 0.8961 & \textcolor{blue}{0.8985} & \textcolor{blue}{0.0206} & \textcolor{blue}{0.9225} & \textcolor{blue}{0.8836} & \textcolor{blue}{0.8233} & \textcolor{blue}{0.8400} & \textcolor{blue}{0.0337} & 453.0M & 139.7G & 29.3\\
\midrule
    DCNet$_{22}$ \cite{tu2022weakly}  & \textcolor{blue}{0.9084} & 0.8599 & 0.7986 & \textcolor{blue}{0.8262} & 0.0364 & \textcolor{red}{\textbf{0.9432}} & 0.9149 & 0.8891 & \textcolor{red}{\textbf{0.9015}} & 0.0233 & 0.9075 & 0.8536 & 0.7899 & 0.8237 & 0.0409 & 24.1M & 207.2G & 14.5\\	
    MROS$_{23}$ \cite{10315195}  & 0.9015 & \textcolor{blue}{0.8751} & \textcolor{blue}{0.8098} & 0.8204 & \textcolor{blue}{0.0361} & 0.9384 & 0.9254 & \textcolor{blue}{0.8971} & 0.8978 & 0.0207 & 0.9092 & 0.8742 & 0.8124 & 0.8281 & 0.0389 & 90.4M & 45.3G & 12.2\\
\midrule
    AlignSal (Ours) & \textcolor{red}{\textbf{0.9144}} & \textcolor{red}{\textbf{0.8880}} & \textcolor{red}{\textbf{0.8175}} & \textcolor{red}{\textbf{0.8304}} & \textcolor{red}{\textbf{0.0317}} & 0.9409 & \textcolor{red}{\textbf{0.9302}} & \textcolor{red}{\textbf{0.8980}} & 0.8965 & \textcolor{red}{\textbf{0.0193}} & \textcolor{red}{\textbf{0.9277}} & \textcolor{red}{\textbf{0.8873}} & \textcolor{red}{\textbf{0.8290}} & \textcolor{red}{\textbf{0.8424}} & \textcolor{red}{\textbf{0.0321}} & 27.1M & 22.9G & 30.8\\
    \bottomrule
    \hline
  \end{tabular}}
\end{table*}
\begin{table*}[!htp]
  \centering
  \fontsize{8}{10}\selectfont
  \renewcommand{\arraystretch}{1.1}
  \renewcommand{\tabcolsep}{0.5mm}
  \scriptsize
  %\captionsetup{labelformat=empty}
  \caption{Quantitative Comparison Between Our AlignSal and Lightweight (L) and Heavyweight (H) BSOD Models on the VT821 \cite{wang2018rgb}, VT1000 \cite{tu2019rgb}, and VT5000 \cite{9767629} Datasets. The Best Results are Labeled \textcolor{red}{\textbf{Red}} and the Second Best Results are Labeled \textcolor{blue}{Blue}. We Tested the FPS of Different Models on an NVIDIA RTX4060 Laptop GPU.}
\label{tab:alignedBSOD}
  \scalebox{1}{
  \begin{tabular}{>{\centering\arraybackslash}p{1.8cm}|c|ccccc|ccccc|ccccc|ccc}
  \hline\toprule
   \multirow{2}{*}{\centering Model} &\multirow{2}{*}{\centering Type} & \multicolumn{5}{c|}{\centering VT821} & \multicolumn{5}{c|}{\centering VT1000} & \multicolumn{5}{c|}{\centering VT5000} \\
    &&$Em\uparrow$ & $Sm\uparrow$ &$wF_{\beta} \uparrow$ & $F_{\beta}\uparrow$ & $\mathcal{M}\downarrow$ & $Em\uparrow$ & $Sm\uparrow$ & $wF_{\beta} \uparrow$ & $F_{\beta}\uparrow$ & $\mathcal{M}\downarrow$ & $Em\uparrow$ & $Sm\uparrow$ & $wF_{\beta} \uparrow$ & $F_{\beta}\uparrow$ & $\mathcal{M}\downarrow$  & Params $\downarrow$ & FLOPs $\downarrow$ & FPS $\uparrow$\\
\midrule	
    CSRNet$_{22}$ \cite{9505635} &L& \textcolor{blue}{0.908} & \textcolor{blue}{0.885} & \textcolor{red}{\textbf{0.821}} & \textcolor{red}{\textbf{0.830}} & 0.038 & 0.925 & 0.918 & 0.878 & 0.877 & 0.024 & 0.905 & 0.868 & 0.797 & 0.811 & 0.042 & 1.0M & 6.3G & 40.2\\	
    OSRNet$_{22}$ \cite{9803225}&L& 0.896 & 0.875 & 0.801 & 0.813 & 0.043 & \textcolor{red}{\textbf{0.935}} & \textcolor{blue}{0.926} & \textcolor{red}{\textbf{0.891}} & \textcolor{red}{\textbf{0.892}} & \textcolor{red}{\textbf{0.022}} & \textcolor{blue}{0.908} & 0.875 & \textcolor{blue}{0.807} & \textcolor{blue}{0.823} & \textcolor{blue}{0.040} & 42.4M & 15.6G & 95.6\\
    LSNet$_{23}$ \cite{10042233} &L& \textcolor{red}{\textbf{0.911}} & 0.878 & 0.809 & \textcolor{blue}{0.825} & \textcolor{red}{\textbf{0.033}} & \textcolor{red}{\textbf{0.935}} & \textcolor{blue}{0.925} & \textcolor{blue}{0.887} & \textcolor{blue}{0.885} & \textcolor{blue}{0.023} & \textcolor{red}{\textbf{0.915}} & \textcolor{blue}{0.877} & 0.806 & \textcolor{red}{\textbf{0.825}} & \textcolor{red}{\textbf{0.037}} & 4.6M & 1.2G & 88.4\\
    \midrule
    AlignSal (Ours) &L& \textcolor{red}{\textbf{0.911}} & \textcolor{red}{\textbf{0.887}} & \textcolor{blue}{0.813} & 0.816 & \textcolor{blue}{0.035} & \textcolor{blue}{0.933} & \textcolor{red}{\textbf{0.932}} & 0.884 & 0.878 & \textcolor{red}{\textbf{0.022}} & \textcolor{red}{\textbf{0.915}} & \textcolor{red}{\textbf{0.884}} & \textcolor{red}{\textbf{0.810}} & 0.819 & \textcolor{red}{\textbf{0.037}} & 27.1M & 22.9G & 30.8\\
\hline\toprule
\hline\toprule
    MIDD$_{21}$ \cite{9454273} &H& 0.895 &0.871 & 0.760 & 0.804 & 0.045 & 0.933 & 0.915 & 0.856 & 0.822 & 0.027 & 0.897 & 0.868 & 0.763 & 0.801 & 0.043 & 52.4M & 216.7G & 86.3\\	
    DCNet$_{22}$ \cite{tu2022weakly}  &H& 0.913 & 0.877 & 0.822 & 0.841 & 0.033 & \textcolor{blue}{0.949} & 0.923 & 0.902 & \textcolor{red}{\textbf{0.911}} & 0.021 & 0.921 & 0.872 & 0.819 & 0.847 & 0.035 & 24.1M & 207.2G & 14.5 \\
    SwinNet$_{22}$ \cite{9611276} &H& \textcolor{blue}{0.926} & 0.904 & 0.818 & 0.847 & 0.030 & 0.947 & \textcolor{blue}{0.938} & 0.894 & 0.896 & 0.018 & \textcolor{blue}{0.942} & \textcolor{red}{\textbf{0.912}} & 0.846 & 0.865 & \textcolor{blue}{0.026} & 198.7M & 124.3G & 25.8\\	
    TNet$_{23}$ \cite{9926193} &H& 0.919 & 0.899 & 0.841 & 0.842 & 0.030 & 0.937 & 0.929 & 0.895 & 0.889 & 0.021 & 0.927 & \textcolor{blue}{0.895} & 0.840 & 0.846 & 0.033 & 87.0M & 39.7G & 59.0 \\
    HRTNet$_{23}$ \cite{9869666} &H& \textcolor{red}{\textbf{0.929}} & \textcolor{blue}{0.906} & \textcolor{blue}{0.849} & \textcolor{blue}{0.853} & \textcolor{blue}{0.026} & 0.945 & \textcolor{blue}{0.938} & \textcolor{blue}{0.913} & 0.900 & \textcolor{blue}{0.017} & \textcolor{red}{\textbf{0.945}} & \textcolor{red}{\textbf{0.912}} & \textcolor{red}{\textbf{0.870}} & \textcolor{red}{\textbf{0.871}} & \textcolor{red}{\textbf{0.025}} & 58.9M & 17.3G & 11.5\\
    MGAI$_{23}$ \cite{10003255} &H& 0.913 & 0.891 & 0.824 & 0.829 & 0.031 & 0.935 & 0.929 & 0.893 & 0.885 & 0.021 & 0.915 & 0.883 & 0.815 & 0.824 & 0.034 & 86.8M & 94.2G & 4.7\\
    WaveNet$_{23}$ \cite{10127616} &H& \textcolor{red}{\textbf{0.929}} & \textcolor{red}{\textbf{0.912}} & \textcolor{red}{\textbf{0.863}} & \textcolor{red}{\textbf{0.857}} & \textcolor{red}{\textbf{0.024}} & \textcolor{red}{\textbf{0.952}} & \textcolor{red}{\textbf{0.945}} & \textcolor{red}{\textbf{0.921}} & \textcolor{blue}{0.909} & \textcolor{red}{\textbf{0.015}} & 0.940 & \textcolor{red}{\textbf{0.912}} & \textcolor{blue}{0.865} & \textcolor{blue}{0.867} & \textcolor{blue}{0.026} & 80.7M & 64.0G & 5.3\\
    LAFB$_{24}$ \cite{wang2024learning} &H& 0.915 & 0.884 & 0.817 & 0.842 & 0.034 & 0.945 & 0.932 & 0.905 & 0.905 & 0.018 & 0.931 & 0.893 & 0.841 & 0.857 & 0.030 & 453.0M & 139.7G & 29.3\\
\midrule
    AlignSal (Ours) &L& 0.911& 0.887 & 0.813 & 0.816 & 0.035 & 0.933 & 0.932 & 0.884 & 0.878 & 0.022 & 0.915 & 0.884 & 0.810 & 0.819 & 0.037 & 27.1M & 22.9G & 30.8\\
    \bottomrule
    \hline
  \end{tabular}}
\end{table*}
\begin{table}[!htp]
\centering
  \fontsize{8}{10}\selectfont
  \renewcommand{\arraystretch}{1}
  \renewcommand{\tabcolsep}{0.7mm}
  \scriptsize
  \caption{Quantitative Comparison of Representatives BSOD Models Before and After Being Equipped with Semantic Contrastive Alignment Loss (SCAL) on the AAV RGB-T 2400 Dataset \cite{10315195}. $*$ Stands for the BSOD Model with SCAL. Subscripted Numbers Indicate Performance \textcolor{red}{Improvement} or \textcolor{blue}{Degradation} Compared to the Original Model.}
  \label{tab:state-of-the-art_w_scal}
  \scalebox{1}{
  \begin{tabular}{c|ccccc}
  \hline\toprule
  Setting &$Em\uparrow$ & $Sm\uparrow$ &$wF_{\beta} \uparrow$ & $F_{\beta}\uparrow$ & $\mathcal{M}\downarrow$\\
\midrule
  MIDD$^{*}$ & 0.8298$_{\emph{\textcolor{blue}{-0.0035}}}$& 0.8593$_{\emph{\textcolor{red}{+0.0069}}}$ & 0.7336$_{\emph{\textcolor{red}{+0.0168}}}$ & 0.6581$_{\emph{\textcolor{red}{+0.0198}}}$ & 0.0103$_{\emph{\textcolor{red}{-0.0011}}}$ \\
  MoADNet$^{*}$  & 0.4744$_{\emph{\textcolor{red}{+0.0117}}}$ & 0.5019$_{\emph{\textcolor{red}{+0.0036}}}$ & 0.0724$_{\emph{\textcolor{red}{+0.0133}}}$ & 0.1147$_{\emph{\textcolor{red}{+0.0162}}}$ & 0.0332$_{\emph{\textcolor{red}{-0.0038}}}$\\
  LSNet$^{*}$ & 0.8948$_{\emph{\textcolor{red}{+0.0021}}}$ & 0.7189$_{\emph{\textcolor{red}{+0.0242}}}$ & 0.3578$_{\emph{\textcolor{red}{+0.0405}}}$ & 0.6795$_{\emph{\textcolor{blue}{-0.0001}}}$ & 0.0330$_{\emph{\textcolor{red}{-0.0081}}}$\\
  MGAI$^{*}$ & 0.8629$_{\emph{\textcolor{red}{+0.0237}}}$ & 0.8471$_{\emph{\textcolor{red}{+0.0175}}}$ & 0.7212$_{\emph{\textcolor{red}{+0.0415}}}$ & 0.6444$_{\emph{\textcolor{red}{+0.0386}}}$ & 0.0113$_{\emph{\textcolor{red}{-0.0010}}}$\\
  CATNet$^{*}$ & 0.7868$_{\emph{\textcolor{red}{+0.0448}}}$ & 0.8412$_{\emph{\textcolor{red}{+0.0275}}}$ & 0.7024$_{\emph{\textcolor{red}{+0.0693}}}$ & 0.5698$_{\emph{\textcolor{red}{+0.0509}}}$ & 0.0155$_{\emph{\textcolor{red}{-0.0027}}}$\\
  \bottomrule
  \hline
  \end{tabular}}
\end{table}
significantly enhancing their performance in handling unaligned bimodal images without taking up additional computational resources during inference.

\emph{3) Effectiveness of SAF:} 
To evaluate the effectiveness of SAF, we replaced it with element-wise addition for bimodal fusion, referred to as ``w/o SAF".
As shown in the $5^{th}$ row of Table \ref{tab:ablation}, 
the performance of AlignSal without SAF significantly declined. 
This finding was further supported by Fig. \ref{fig:attn} and \ref{fig:ablation}. 
In Fig. \ref{fig:attn} (g), 
the model without SAF failed to align RGB and thermal features in the fused features and did not effectively integrate the bimodal complementary information, resulting in confused and overlapping salient regions. 
These issues were particularly noticeable in high-resolution fused feature maps. 
In Fig. \ref{fig:ablation}, the model without SAF failed to capture detailed features (\textit{e.g.}, the small human figures in the $1^{st}$ row) and to discriminate inter-modal interference information (\textit{e.g.}, the statue in the $3^{rd}$ row). 
Moreover, we observed that the full model maintained consistent Params with only 0.1G higher FLOPs compared to the model without SAF. 
This indicates that the complexity of our proposed SAF is comparable to that of element-wise addition.

\subsection{Generalization Experiment}
To further validate the model's generalization, we compared our AlignSal with existing state-of-the-art bimodal dense prediction models across three different tasks: weakly aligned bimodal salient object detection (BSOD), aligned BSOD, and remote sensing change detection (RSCD). All prediction maps used for
\begin{table}[!htp]
\centering
  \fontsize{8}{10}\selectfont
  \renewcommand{\arraystretch}{1}
  \renewcommand{\tabcolsep}{0.8mm}
  \scriptsize
  \caption{Quantitative Comparison Between Our AlignSal and Three Existing RSCD Models on the LEVIR-CD+ Dataset \cite{chen2020spatial}. The Best Results are Labeled \textcolor{red}{\textbf{Red}}.}
  \label{tab:CD}
  \scalebox{1}{
  \begin{tabular}{c|ccccc|cc}
  \hline\toprule
  Model &Rec.$\uparrow$ & Pre. $\uparrow$ & OA $\uparrow$& F1$\uparrow$ & IoU$\uparrow$ & Params $\downarrow$ & FLOPs $\downarrow$\\
\midrule
  FC-EF$_{18}$ \cite{daudt2018fully} &71.77 &69.12 &97.54 &70.42 &54.34 &1.4M &14.1G \\
  FC-Siam-Diff$_{18}$ \cite{daudt2018fully} &74.02 &\textcolor{red}{\textbf{81.49}} &98.26 &77.57 &63.36 &1.4M &18.7G \\
  SNUNet$_{22}$ \cite{fang2021snunet} &\textcolor{red}{\textbf{78.73}} &71.07 &97.83 &74.70  &59.62 &10.2M &176.4G \\
  HANet$_{23}$ \cite{han2023hanet} &75.53 &79.70 &98.22 &77.56 &63.34 &2.6M &70.7G \\
  \midrule
  AlignSal (Ours) &77.78 &79.12 & \textcolor{red}{\textbf{98.45}} & \textcolor{red}{\textbf{78.44}} &\textcolor{red}{\textbf{64.53}} & 27.1M & 22.9G \\
  \bottomrule
  \hline
  \end{tabular}}
\end{table}
\begin{figure}[!htp]
	\centering \includegraphics[width=0.40\textwidth]{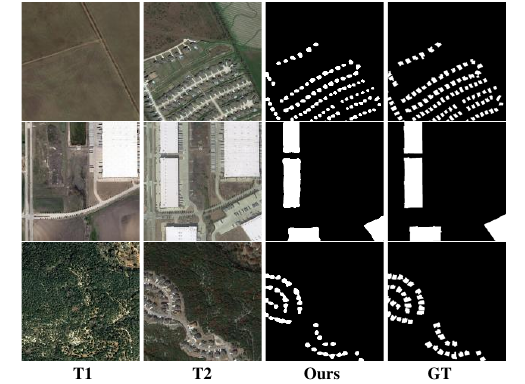}
	\caption{Visual results of our AlignSal on the LEVIR-CD+ dataset. T1 and T2 stand for pre-event and post-event images, respectively. GT is ground truth.}
\label{fig:CD}
\end{figure}
evaluation were obtained from the authors or generated using their open-source codes.

\emph{1) Weakly Aligned BSOD:} We first compared our AlignSal with existing BSOD models, including
MIDD \cite{9454273}, MoADNet \cite{jin2022moadnet}, DCNet \cite{tu2022weakly}, OSRNet \cite{9803225}, LSNet \cite{10042233}, TNet \cite{9926193}, MGAI \cite{10003255}, CATNet \cite{sun2023catnet}, PICRNet \cite{cong2023point}, LAFB \cite{wang2024learning}, and MROS \cite{10315195} on three unaligned datasets, namely unaligned-VT821, unaligned-VT1000, and unaligned-VT5000 \cite{tu2022weakly}. 

As shown in Table \ref{tab:generalization}, AlignSal achieved a significant lead with lower computational cost and faster inference speed compared to the current top-performing models, such as MROS \cite{10315195}, DCNet \cite{tu2022weakly}, and LAFB \cite{wang2024learning}. 
These comprehensive experiments confirmed that AlignSal, with our proposed key components, can generalize well to different unaligned bimodal datasets.

\emph{2) Aligned BSOD:} We then compared our AlignSal with three lightweight BSOD models, namely CSRNet \cite{9505635}, OSRNet \cite{9803225}, and LSNet \cite{10042233}, and eight heavyweight BSOD models, namely MIDD \cite{9454273}, DCNet \cite{tu2022weakly}, SwinNet \cite{9611276}, TNet \cite{9926193}, HRTNet \cite{9869666}, MGAI \cite{10003255}, WaveNet \cite{10127616}, and LAFB \cite{wang2024learning} on the VT821 \cite{wang2018rgb}, VT1000 \cite{tu2019rgb}, and VT5000 \cite{9767629} datasets.

As shown in Table \ref{tab:alignedBSOD}, AlignSal exhibited good generalization capabilities on the three RGB-thermal BSOD datasets, with performance comparable to existing lightweight models.
Although AlignSal's performance lags behind most heavyweight models, this result is expected.
Heavyweight models are designed with specialized and complex structures that can fully exploit the deep relationships between alignment features in manually aligned data.
\begin{figure}[!htp]
	\centering \includegraphics[width=0.48\textwidth]{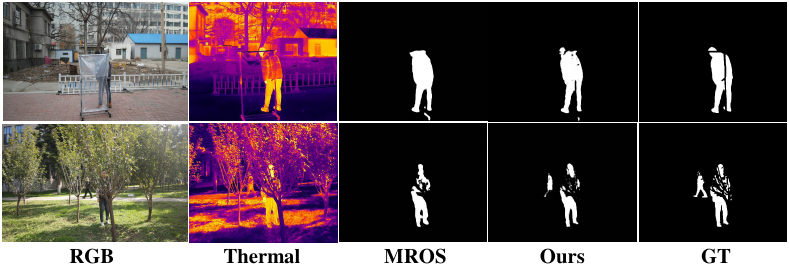}
	\caption{Examples of failure cases of our AlignSal and MROS \cite{10315195} dealing with object occlusion scenes. GT is ground truth.}
\label{fig:failure}
\end{figure}
In contrast, AlignSal is designed for unaligned data with a lightweight structure, which is more suitable for real-world detection scenarios where data alignment may not always be feasible.
For example, in applications such as industrial inspection and autonomous driving, inter-modal spatial misalignment can significantly degrade the performance of models that rely on precise alignment (as shown in Table \ref{tab:comp_AAV}).
Furthermore, the need for lightweight deployment in these scenarios makes heavyweight models impractical,
highlighting the importance of AlignSal's design philosophy.

\emph{3) RSCD:} To evaluate the generalizability of our AlignSal in high-altitude scenarios, we compared it with three existing RSCD models, including FC \cite{daudt2018fully}, SNUNet \cite{fang2021snunet}, and HANet \cite{han2023hanet} on the LEVIR-CD+ dataset \cite{chen2020spatial}. 

As shown in Table \ref{tab:CD}, our AlignSal achieves competitive performance compared to three state-of-the-art RSCD models. Notably, although AlignSal is designed for unaligned data, its components - SCAL and SAF - enable effective processing of spatiotemporal features. SCAL facilitates the mutual refinement of two-stream spatiotemporal features by contrastive communication, while SAF effectively identifies regions of change between the two input images. In addition, visual results in Fig. \ref{fig:CD} demonstrated AlignSal's excellent generalization capabilities, allowing it to capture tiny changes even in high-altitude scenarios.

\subsection{Failure Cases}
Despite the impressive performance of our AlignSal for unaligned BSOD, there are still some challenging scenarios in which it does not produce satisfactory results. 
As shown in Fig. \ref{fig:failure}, 
when salient objects are occluded, 
both our model and the current top-performing MROS \cite{10315195} struggle to accurately distinguish the boundary between the occluder and the salient objects. Future improvements, therefore, could focus on developing more advanced multi-dimensional filtering mechanisms to enhance scene understanding, 
particularly in cases where the occluder is a fine-grained structure.

\section{Conclusions}
In this paper, we examined the relationship between unaligned modalities and proposed AlignSal, which achieves both real-time performance and high accuracy for AAV-based unaligned BSOD. 
To this end, we first designed SCAL to align bimodal semantic features in a contrastive manner and enhance inter-modal information exchange. 
Notably, SCAL not only works on our proposed AlignSal, but also improves the performance of various existing aligned BSOD models on the AAV RGB-T 2400 dataset. 
Second, we proposed the FFT-based SAF to align slightly offset features in multiple dimensions and facilitate bimodal fusion by efficiently acquiring global relevance. 
Extensive experiments demonstrated that AlignSal achieves significantly faster inference speed and superior performance compared to existing models on the AAV RGB-T 2400 dataset, three weakly aligned BSOD datasets, three aligned BSOD datasets, and one remote sensing change detection dataset. 
These advantages make AlignSal suitable for deployment on AAVs in practice,
enabling its use in industrial and real-life applications that require both timeliness and accuracy.

\bibliographystyle{IEEEtran}
\balance
\bibliography{ref}

\end{document}